\documentclass[11pt,onecolumn]{IEEEtran}
\usepackage{epsfig}
\usepackage{subfigure}
\usepackage{amssymb}
\usepackage{amsbsy}
\usepackage{amsmath}
\usepackage{cite}
\usepackage{url}
\usepackage{cite}
\usepackage{color}
\usepackage{float}
\usepackage{times,amsmath,epsfig}
\usepackage{xspace,latexsym,syntonly}
\usepackage{amssymb}
\usepackage{amsfonts}
\usepackage{textcomp}
\usepackage{subfigure}
\usepackage{amsbsy}
\usepackage[ruled]{algorithm2e}
\usepackage{algpseudocode}
\usepackage{hyperref}

\usepackage{array}
\usepackage{makecell}

\usepackage{booktabs}

\usepackage{geometry}

\usepackage{cite} 

\usepackage{tikz}

\newtheorem{theorem}{Theorem}
\newtheorem{lemma}{Lemma}

\newtheorem{definition}{Definition}

\newcommand{\beq}{\begin{equation}}
\newcommand{\eeq}{\end{equation}}
\newcommand{\bea}{\begin{array}}
\newcommand{\ena}{\end{array}}
\newcommand{\bds}{\begin {itemize}}
\newcommand{\eds}{\end {itemize}}
\newcommand{\bdf}{\begin{definition}}
\newcommand{\blm}{\begin{lemma}}
\newcommand{\edf}{\end{definition}}
\newcommand{\elm}{\end{lemma}}
\newcommand{\bthm}{\begin{theorem}}
\newcommand{\ethm}{\end{theorem}}
\newcommand{\bprp}{\begin{prop}}
\newcommand{\eprp}{\end{prop}}
\newcommand{\bcl}{\begin{claim}}
\newcommand{\ecl}{\end{claim}}
\newcommand{\bcr}{\begin{coro}}
\newcommand{\ecr}{\end{coro}}
\newcommand{\bquest}{\begin{question}}
\newcommand{\equest}{\end{question}}

\newcommand{\larrow}{{\larrow}}

\def\urltilda{\kern -.15em\lower .7ex\hbox{\~{}}\kern .04em}

\renewcommand{\baselinestretch}{1.5}

\begin{document}
\title{Control-Optimized Deep Reinforcement Learning for Artificially Intelligent Autonomous Systems}

\author{Oren Fivel$^1$, Matan Rudman$^1$, Kobi Cohen$^1$
\thanks{$^1$School of Electrical and Computer Engineering, Ben-Gurion University of the Negev, Beer-Sheva, Israel (Email:fivel@post.bgu.ac.il; rudman@post.bgu.ac.il; yakovsec@bgu.ac.il).
	}
\thanks{Open-source code for the algorithm and simulations developed in this paper can be found on GitHub at \cite{web:OrenGit}.}
}


\date{}
\maketitle

\begin{abstract}
\label{sec:abstract}
Deep reinforcement learning has become a powerful tool for complex decision-making in machine learning and AI. However, traditional methods often assume perfect action execution, overlooking the uncertainties and deviations between an agent’s selected actions and the actual system response. In real-world engineering applications—such as robotics, mechatronics, and communication networks—execution mismatches arising from system dynamics, hardware constraints, and latency can significantly degrade performance.

This work advances AI by developing a novel control-optimized deep reinforcement learning framework that explicitly models and compensates for action execution mismatches, a challenge largely overlooked in existing methods. Our approach establishes a structured two-stage process: determining the desired action (e.g., force or torque) and selecting the appropriate control signal (e.g., voltage) to ensure proper execution. It trains the agent while accounting for action mismatches and controller corrections. By incorporating these factors into the training process, the AI agent optimizes the desired action with respect to both the actual control signal and the intended outcome, explicitly considering execution errors. This approach enhances robustness, ensuring that decision-making remains effective under real-world uncertainties. 

Our approach offers a substantial advancement for engineering practice by bridging the gap between idealized learning and real-world implementation. It equips intelligent agents operating in engineering environments with the ability to anticipate and adjust for actuation errors and system disturbances during training. We evaluate the framework in five widely used open-source mechanical simulation environments we restructured and developed to reflect real-world operating conditions, showcasing its robustness against uncertainties and offering a highly practical and efficient solution for control-oriented applications.
\vspace{0.5cm}
\end{abstract}

\def\keywords{\vspace{.5em}
{\bfseries\textit{Keywords}---\,\relax%
}}
\def\endkeywords{\par}
\keywords
artificial intelligence, autonomous systems, decision making, deep reinforcement learning, optimization and control.

\begin{table}[h]
\centering
\caption{List of Abbreviations}
\begin{tabular}{@{}lll@{}}
\toprule
Abbreviation &  & Description                                   \\ \midrule
AI         &  & Artificial Intelligence             
      \\
BEMF         &  & Back Electromagnetic Force                    \\
CO-DRL       &  & Control-Optimized Deep Reinforcement Learning \\
D3QN         &  & Dueling Double Deep Q-Network   
       \\
DDPG         &  & Deep Deterministic Policy Gradient  
        \\
DNN          &  & Deep Neural Network         
\\
DQN          &  & Deep Q-Network                                \\
DRL          &  & Deep Reinforcement Learning                   \\
IRL          &  & Integral Reinforcement Learning  
        \\
KLD          &  & Kullback–Leibler Divergence
        \\
LQR          &  & Linear Quadratic Regulator                    \\
LQT          &  & Linear Quadratic Tracker 
      \\
MAB           &  & Multi-Armed Bandit                               \\
NN           &  & Neural Network                                \\
PD           &  & Proportional-Derivative                       \\
PI           &  & Proportional-Integral                         \\
PID          &  & Proportional-Integral-Derivative              \\
PPO          &  & Proximal Policy   Optimization                \\
ReLU         &  & Rectified Linear Unit                         \\
RL           &  & Reinforcement Learning      
        \\
SARSA        &   & State–Action–Reward–State–Action

\\ \bottomrule
\end{tabular}
\end{table}

\section{Introduction}
\label{ssec:intro}

Deep Reinforcement Learning (DRL) has become a cornerstone in the field of AI, particularly for solving complex decision-making problems in dynamic and uncertain environments. By enabling agents to learn optimal behaviors through interaction with their surroundings, DRL has shown remarkable success across various applications, including robotics, autonomous vehicles, and mechatronic systems. These systems often operate under uncertain conditions, where the link between an agent’s decisions and the resulting actions can be influenced by factors such as delays, noise, and system dynamics. This makes DRL an ideal approach for tasks requiring adaptive and resilient decision-making. As the demand for intelligent, autonomous systems continues to grow, the ability of DRL to handle real-world complexities makes it a powerful tool for advancing the next generation of autonomous technologies. In this paper, we focus on addressing the challenge of uncertainty in action execution by modeling the dynamics of the agent’s actuation system, improving the alignment between desired and actual actions in real-world systems.

\subsection{Contributions}
\label{ssec:contributions}

To address the challenge of uncertain and imperfect action execution in real-world AI-driven autonomous systems, we propose Control-Optimized Deep Reinforcement Learning (CO-DRL), a novel DRL framework that integrates control theory to enhance decision-making and execution. By incorporating a feedback loop control mechanism, CO-DRL ensures that intended actions are effectively carried out despite inner system dynamics, disturbances, and model uncertainties. CO-DRL distinguishes between determining the desired action (e.g., force/torque) and selecting the control signal (e.g., voltage) needed to achieve it. By incorporating a feedback loop control mechanism, our approach optimizes both decision-making and execution, ensuring consistency between intended and realized actions.

Our framework extends OpenAI Gym’s classic control environments by incorporating a DC motor model with a Proportional-Integral-Derivative (PID) controller for action tracking. This modular implementation allows flexible customization of the actuation system, control architecture, and DRL algorithms, enabling a seamless combination of decision-making and execution strategies. To support both research and engineering applications, we release an open-source Python implementation of CO-DRL, available on GitHub (see \cite{web:OrenGit}). The software is designed with object-oriented principles, facilitating integration into a variety of control-oriented AI systems.

For researchers and developers, CO-DRL offers a practical toolset for improving the robustness and reliability of DRL-based autonomous systems. By explicitly modeling and compensating for execution mismatches, it enables more stable and predictable policy behavior in complex, real-world settings. The provided environments serve as a testbed for developers and engineers to evaluate and refine DRL policies under realistic conditions, supporting applications in robotics, industrial automation, and other domains where execution fidelity is critical.

We evaluate CO-DRL across five widely used OpenAI Gym environments—Acrobot, CartPole, Mountain Car, Continuous Mountain Car, and Pendulum—chosen for their relevance to classic control tasks and their representativeness of common challenges in engineering systems. To demonstrate the generality and adaptability of our framework, we integrate it with a range of DRL algorithms, including Proximal Policy Optimization (PPO) \cite{schulman2017proximal}, Deep Q-Networks (DQN) \cite{mnih2013playing}, Tile Coding \cite{sutton1995generalization}, and Deep Deterministic Policy Gradient (DDPG) \cite{lillicrap2015continuous}. Experimental results show that CO-DRL consistently improves performance under conditions of actuation uncertainty and system disturbance. These findings highlight its practical utility and robustness, making it a strong candidate for deployment in AI-driven autonomous systems within engineering domains such as robotics, control systems, and automation.

\subsection{Related Work}
\label{ssec:related}

RL and particularly DRL algorithms have been implemented as part of a control law design for various autonomous dynamic systems, such as the pendulum (including cartpole and acrobot) \cite{puriel2018reinforcement, lim2020federated,  livne2020pops, dao2021adaptiveRL_pendulum}, intelligent transportation and autonomous vehicles \cite{haydari2022deep, guangwen2024achieving}, drones and quadrotors \cite{azar2021drone, hwangbo2017controlQuadrotor}, and robotic systems \cite{kartoun2010human-robot, taitler2017learning, elyasaf2019usingBPRoboCup}. These methods have been increasingly integrated into control law design for autonomous systems, with frameworks such as Associative Search Element, Adaptive Critic Element \cite{barto_sutton_anderson_1983}, Tile Coding \cite{sutton1995generalization}, Deep Q-Networks (DQN) \cite{li2019solve}, and DRL-based proportional-derivative (PD) control \cite{puriel2018reinforcement}. In recent studies, DRL-based methods continue to push the envelope in areas like multi-agent systems and optimization tasks. These DRL frameworks generally optimize decision-making by maximizing rewards, assuming perfect action execution. However, real-world systems, such as those with actuators (e.g., motors), introduce discrepancies between desired and actual actions due to inner dynamics, disturbances, and model uncertainties, as considered in this paper. 

Classical RL methods, such as multi-armed bandit (MAB) and tabular Q-learning, have long been used in engineering to solve decision-making problems under uncertainty in diverse areas, such as biomedical engineering, financial investment, robotics and mechanic systems, and communication networks
\cite{gafni2020learning, gafni2022learning, amar2023online, busacca2024mad, gafni2022distributed, busacca2025distributed, ami2025client}. Moreover, reinforcement learning has been applied to other specific classic control problems. For example, \cite{xue2020off} applies off-policy integral reinforcement learning (IRL) to solve optimal linear quadratic tracker (LQT) problems for continuous-time two-time-scale processes, addressing both slow and fast states with linear quadratic regulator (LQR) and LQT. In  \cite{peng2020reinforcement}, the authors investigate an online reinforcement Q-learning algorithm to design a model-free $H_\infty$ tracing controller for unknown discrete-time linear systems, which is demonstrated in a simulation of a single-phase voltage-source UPS inverter. Additionally, \cite{mukherjee2021reduced} explores a reduced-dimensional reinforcement learning approach for linear time-invariant singularly perturbed systems, focusing on the separation of slow and fast states, with the action determined by LQR control subject to the slow states. 

While classic approaches are effective mainly in structured environments with low-dimensional state and action spaces, as engineering systems have grown in complexity, DRL has emerged as a powerful extension, combining the strengths of RL with deep learning to handle complex, unstructured problems. This advancement has led to significant progress across a wide range of engineering domains. In AI-based communication networks, DRL has been employed for dynamic spectrum access, resource management, and cognitive radio \cite{naparstek2019deep, ZhongGursoy2020DRLEdgeCachWirelessNet, gahtan2023using, abbasi2021deep, bokobza2023deep, paul2023multi, bai2025deep, LiFang2018IntelligentDRL, cohen2025sinr}. In the context of computer vision, it has been used for 
image categorization \cite{yerramreddy2024harnessing}, automated classification \cite{dubayan2025automated}, and intelligent expert-aided classification \cite{puzanov2018deep, puzanov2020deep}. Additionally, DRL methods have been adopted for AI-based anomaly detection and active hypothesis testing \cite{kartik2018policy, ZhongGursoy2019DACRLAnomalyDetec, szostak2022decentralized, stamatelis2023deep, szostak2024deep, kalouptsidis2025neural}, as well as health-related applications that require adaptive and robust decision-making \cite{puzanov2020deep, liu2025environment, dubayan2025automated}. These diverse applications demonstrate the adaptability of DRL in solving complex, real-world problems beyond classical control, further highlighting its growing role in modern engineering systems.

However, despite these advancements, existing DRL-based controllers often lack explicit mechanisms to track and correct execution mismatches caused by actuator inner dynamics and external disturbances. In this work, we address this gap by developing CO-DRL, a novel framework that optimizes both decision-making and execution during training. This approach ensures consistency between intended and realized actions, enhancing robust decision-making and execution in real-world autonomous systems.

\section{Methodology} 

We begin with a formal introduction of the system model and problem statement, followed by a presentation of the proposed control-optimized DRL framework designed to address the problem.

\subsection{System Model and Problem Statement} 

In a mechanical system environment, the state space $\mathcal{S}$ consists of the continuous set of positions (both linear and angular) and their corresponding velocities for the bodies and particles within the system, determined by its degrees of freedom. Let $s_t \in \mathcal{S}$ represent the system's state at discrete time $t$.  

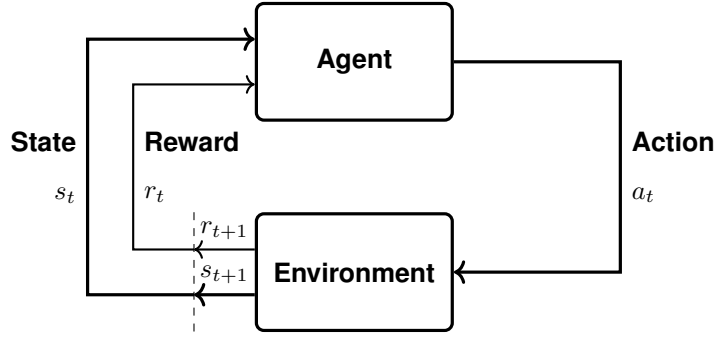
\begin{figure}[H]
    \centering    
    \usetikzlibrary{shapes,arrows.meta,shadows,positioning}

\tikzset{
  frame/.style={
    rectangle, draw,
    text width=6em, text centered,
    minimum height=4em,fill=white,
    rounded corners=1mm,
  },
  line/.style={
    draw, -{Latex},rounded corners=0mm,
  }
}

\begin{tikzpicture}[font=\small\sffamily\bfseries,very thick,node distance = 4cm]
\node [frame] (agent) {Agent};
\node [frame, below=1.2cm of agent] (environment) {Environment};
\draw[->] (agent) -- ++ (3.5,0) |- (environment)
node[right,pos=0.25,align=left] {Action\\ $a_t$};
\coordinate[left=8mm of environment] (P);
\coordinate[above=3mm of environment.west] (ENW);
\coordinate[below=3mm of environment.west] (ESW);
\coordinate[above=3mm of agent.west] (ANW);
\coordinate[below=3mm of agent.west] (ASW);
\draw[thin,dashed] (P|-environment.north) -- (P|-environment.south);
\draw[->] (ESW) -- (P |- ESW)
node[midway,above]{$s_{t+1}$};
\draw[->,thick] (ENW) -- (P |- ENW)
node[midway,above]{$r_{t+1}$};
\draw[->] (P |- ESW) -- ++ (-1.4,0) |- (ANW)
node[left, pos=0.25, align=right] {State\\ $s_t$};
\draw[->,thick] (P |- ENW) -- ++ (-0.8,0) |- (ASW)
node[right,pos=0.25,align=left] {Reward\\ $r_t$};
\end{tikzpicture}
    \caption{An illustration of a basic reinforcement learning framework.}
    \label{fig:basicRL}
\end{figure}

An agent in the mechanical system corresponds to the actuation component that applies external forces or torques to certain bodies within the system. Let $\mathcal{A}$ denote the action space available to the agent, with $a_t \in \mathcal{A}$ representing an action taken at time $t$. The action space $\mathcal{A}$ can be either continuous or discrete. Fig. \ref{fig:basicRL} illustrates a basic RL framework.

At each iteration, the state $s_t$ and reward $r_t$ are provided to the Agent block. The agent selects an action aimed at maximizing the future accumulated reward. This action is then passed to the Environment block, which generates the next state $s_{t+1}$ and the corresponding reward $r_{t+1}$. 

In practical mechanical systems, each action $a_t$ is mapped to an external force/torque as follows:
 \begin{equation}\label{eq:Action2Force}
    F_t = \texttt{ActionFeature}(a_t)=F{\{a_t\}}, 
 \end{equation}
where \texttt{ActionFeature}$(\cdot)$ (or $F{\{\cdot\}}$ for brevity) is the function that maps an action $a_t$ to an external force/torque $F_t$. For a continuous action space, it is often convenient to define the mapping as a simple linear relationship, where the action directly represents the applied force/torque: $F_t = a_t$. In the discrete case, it is often convenient to define the action space as a finite set of indexed actions $\{0,1,\dots,N-1\}$, where each index $k$ corresponds to a predefined external force/torque value $F{\{k\}} \in \mathbb{R}$.

At each time $t$, a new state $s_{t+1}$ is generated by 
\begin{equation}\label{eq: state update}
    s_{t+1}=\texttt{EnvStateUpdate}(s_t,F_t),
\end{equation}
where $\texttt{EnvStateUpdate}(s_t, F_t)$ is a function of the current state $s_t$ and the force/torque $F_t = F{\{a_t\}}$ associated with the current action $a_t$. Let $p_t$ and $v_t$ denote the position and velocity of the mechanical object, respectively. Applying Newton's Second Law, a typical next-state update is performed as follows:
\begin{align}
    v_{t+1} &= v_t+\texttt{Acceleration}(s_t,F_t) {\Delta t},   \label{eq:V}\\
    p_{t+1} &= p_t+v_t {\Delta t}   \label{eq:P},
\end{align}
where \texttt{Acceleration}($\cdot$,$\cdot$) is a function of the current state $s_t$ and the force/torque $F_t$, representing the acceleration of the bodies, and ${\Delta t}$ is the integration time step. The following integration methods are used in the five openAI Gym environments analyzed in this work. In MountainCar (both discrete and continuous), the next position $p_{t+1}$ is updated using semi-implicit Euler integration, meaning it is computed based on the next velocity $v_{t+1}$ rather than the current velocity $v_t$. In Pendulum, the next position is updated via regular Euler integration, using the current velocity $v_t$. In CartPole, either regular Euler integration or semi-implicit Euler integration can be selected. In Acrobot, integration is performed using the Runge-Kutta method.

After the state update, the next reward $r_{t+1}$ is updated as follows: 
\begin{equation}\label{eq: reward update}
    r_{t+1} = \texttt{Reward}(s_{t+1},s_t, F_t),
\end{equation}
where \texttt{Reward} is a function of the updated state $s_{t+1}$, the pre-update state $s_t$, and the applied force/torque $F_t$ corresponding to the action. The reward function can take various forms, such as quadratic or boolean. The next observations $o_{t+1}$ are given by: 
\begin{equation}\label{eq: obs update}
    o_{t+1} = \texttt{getObs}(s_{t+1},s_t, F_t).
\end{equation}
In general, the observations $o_t$ and the states $s_t$ are not identical. For instance, in Pendulum, the state consists of the pendulum's angle and angular velocity (dimension $= 2$), whereas the observation represents the angle using its sine and cosine components instead (dimension $= 3$).

The objective of the algorithm is to enable an agent to make effective sequential decisions in a dynamic environment. The DRL algorithm receives observations of the environment and, in some cases, the underlying state, along with the corresponding rewards. Based on this information, the algorithm outputs an action to be executed. The goal is to solve the decision-making problem by learning an optimal policy that adapts to the environment’s dynamics and selects actions that maximize cumulative rewards over time. The learning is done by training a deep neural network (DNN) to approximate the value function or the policy, which is updated iteratively based on the agent's interactions with the environment, allowing it to improve decision-making over time.

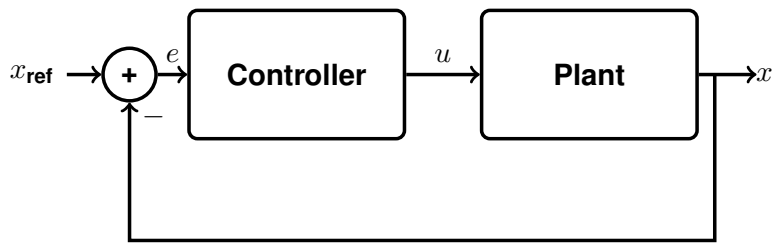
\begin{figure}[H]
    \centering
    \scalebox{1.1}{\usetikzlibrary{shapes,arrows.meta,shadows,positioning}

\tikzset{
  frame/.style={
    rectangle, draw,
    text width=6em, text centered,
    minimum height=4em,fill=white,
    rounded corners=1mm,
  },
  line/.style={
    draw, -{Latex},rounded corners=1mm,
  }
}

\tikzstyle{sum} = [draw, circle, node distance=1cm]

\begin{tikzpicture}[font=\small\sffamily\bfseries,very thick,node distance = 4cm]
\node[frame](controller) {Controller};
\node [frame, right of=controller, node distance=3.5cm] (plant) {Plant};
\node[sum, left of=controller, node distance = 2cm] (sum) {+};
\draw[->] (controller)  -- node[above, pos = 0.5] {$u$} (plant) ;
\draw[->] (sum) -- node[above, pos = 0.5] {$e$} (controller);
\draw[->] (plant) -- (5.5,0) node[left,pos=1.5,align=left] {$x$};
\draw[->] (5,0) -- (5,-2) -- (-2, -2) -- (sum) node[right,pos=0.9,align=right] {$-$};
\coordinate [left of=sum, node distance = 0.75cm] (xref);
\draw[->]  (xref) [left] node{$x_{\text{ref}}$}--(sum) ;

\end{tikzpicture}}
    \caption{An illustration of a reinforcement learning framework with closed-loop tracking control considered in this paper.}
    \label{Fig Basic Closed Loop Tracking Control}
\end{figure}

Unfortunately, the design and implementation of DRL algorithms for applying the optimal force/torque as an intended action in Gym's classic control environments do not account for the agent's uncertainty caused by the actuator's internal dynamics. Our approach is to generalize the action mechanism so that the force/torque in \eqref{eq:Action2Force} incorporates the uncertainty introduced by the inner dynamics of an actuation system, such as a DC motor. Additionally, we extend the DRL block diagram in Fig. \ref{fig:basicRL} to be based on a closed-loop tracking control framework, as illustrated Fig. \ref{Fig Basic Closed Loop Tracking Control}.

In control theory notation, the state vector $x$ is fed back and compared with a reference signal $x_\text{ref}$. The resulting error, $e = x_\text{ref} - x$, is input to the Controller block (analogous to the Agent), which generates a control signal $u$ (corresponding to the Action). This control signal $u$ is then applied to the Plant block (analogous to the Environment), producing a new state $x$, and the loop continues. 

\subsection{The Control-Optimized Deep Reinforcement Learning (CO-DRL) Framework for AI-Driven Autonomous Systems}

We model the agent's uncertainty by defining the force/torque in \eqref{eq:Action2Force} as the \emph{desired} force/torque rather than the actual one. This desired force/torque, along with the actuated body's velocity, is input into a subsystem comprising a DC motor electrical circuit and a PID controller.

\subsubsection{Integration of DC Motor and PID Controller into the CO-DRL Framework}

For simplicity and as a proof of concept, we assume a linear relationship between force and electrical current, as well as between the voltage drop across the DC motor (the BEMF) and the velocity of the body (in $m/sec$). These assumptions are motivated by fundamental mechanical relationships:
\begin{align}
    \tau &= r \times F, \label{eq: torqe WRT force}\\
    v &= \omega \times r, \label{eq: vel WRT angle vel} 
\end{align}
where $\tau$ represents torque, $F$ denotes force, $v$ is the linear velocity, $\omega$ is the angular velocity, and $r$ is the arm parameter or the rotation radius. In the DC motor model, the following linear relationships hold:
\begin{align}
    \tau &= k_T i, \label{eq: torqe WRT current}\\
    V_{BEMF} &= k_E \omega, \label{eq: Voltage WRT angle vel} 
\end{align}
where $i$ represents the electrical current fed into the DC motor, $V_{BEMF}$ is the voltage drop on the DC motor, and $k_T$ and $k_E$ are the torque constant and the BEMF constant, respectively. The discrete time version of the electrical current $i_t$ in a DC motor is given by:
\begin{equation}\label{eq: current update}
    i_{t+1} = i_t + \frac{u_t-V^{BEMF}_t-Ri_t}{L} {\Delta t}
\end{equation}
where ${\Delta t}$ denotes the integration time, $u_t$ the voltage control input, $R$ is the resistor, and $L$ is the inductor. The voltage control input $u_t$ is constructed by a PID controller subject to the electrical current error (defined later in \eqref{eq: current error}). A typical PID controller in discrete time has the following structure:
\begin{equation} \label{eq: PID control signal}
   u_t=k_P e_t + k_I I_{e_t} + k_D D_{e_t}, 
\end{equation}
where 
\begin{align}
    &I_{e_t}=I_{e_{t-1}}+e_{t}{\Delta t},\label{eq: Integral in PID} \\
    &D_{e_t}= \frac{e_t-e_{t-1}}{\Delta t}.\label{eq: derivative in PID}
\end{align}
Here, $I_{e_t}$ represents the recursive approximation of the integral of the error, $D_{e_t}$ is the approximation of the error's derivative, and $k_P$, $k_I$, and $k_D$ are the PID gains for the proportional, integral, and derivative terms, respectively. However, we modify the standard PID control structure by introducing the $k_E\omega$ term to compensate for the BEMF, enabling pre-analysis of the DC electrical circuit without a motor. Additionally, we apply a saturation to the control signal, i.e.,

\begin{equation}\label{eq: PID control signal with BEMF and SAT}
  \begin{split}
  &u^\textrm{temp}_t = k_P e_t + k_I I_{e_t} + k_D D_{e_t} + k_E\omega, \\
    &u_t=\text{sat}(u^\textrm{temp}_t,-u_\text{max},u_\text{max}),
\end{split} 
\end{equation}
where,
\begin{equation*}
\text{sat}(u,u_\text{min},u_\text{max})=
    \begin{cases}
    u_\text{min}, & u<u_\text{min},\\
    u, & u_\text{min}\leq u \leq u_\text{max},\\
    u_\text{max},& u>u_\text{max}.
    \end{cases}
\end{equation*}

For each input $u_t$ in \eqref{eq: PID control signal}, a new electrical current $i_{t+1}$ is calculated using \eqref{eq: current update}, which generates the actual torque 
\eqref{eq: torqe WRT current} on the mechanical system. Thus, the actual torque (and force) is assumed to be linearly related to the subsequent electrical current:
 \begin{equation}\label{eq: actual force WRT actual current}
    F_{t+1} = k_T i_{t+1}.
 \end{equation}

\subsubsection{Architecture of the Complete CO-DRL Framework for Autonomous Systems}
\label{ssec: Architecture of the Modified System Model}

We now introduce the complete CO-DRL framework for autonomous systems. To define the agent's actions in terms of a desired force/torque, we reformulate \eqref{eq:Action2Force} as follows:
 \begin{equation}\label{eq:Action2desiredForce}
    F^\text{desired}_t = \texttt{ActionFeature}(a_t)=F^\text{desired}{\{a_t\}},
 \end{equation}
where the superscript $\emph{desired}$ refers to a desired force/torque.
Using \eqref{eq: torqe WRT current}, we compute the desired electrical current as a reference value
 \begin{equation}\label{eq: desired force 2 ref current}
    i^{\text{ref}}_t = F^\text{desired}_t/k_T,
 \end{equation}
 and the error
 \begin{equation}\label{eq: current error}
    e_t = i^{\text{ref}}_t-i_t.
 \end{equation}
To construct the electrical model of the DC motor, we extract the velocity of the actuated body from the current state $s_t$ as it corresponds to the motor speed. Let
\begin{equation}\label{eq: Actuated Velocity}
    w_t = \texttt{ActuatedVelocity}(s_t)
 \end{equation}
represent the velocity of the actuated body. The error $e_t$ and the actuated velocity $w_t$ are fed into a DC motor sub-environment, which then generates the next actual force/torque based on \eqref{eq: current update}, \eqref{eq: Integral in PID}-\eqref{eq: actual force WRT actual current}:
\begin{equation}\label{eq: actual force}
    F_{t+1} = \texttt{DCMotorENVStep}(e_t, w_t),
\end{equation}
and this computed version of the actual force/torque $F_{t+1}$  is then used in the state update equation \eqref{eq: state update}. Notably, in this formulation, $F_{t+1}$ depends on $a_t$ rather than $F_t$.

\begin{figure}[H]
    \centering
    \subfigure[An illustration of the proposed CO-DRL framework.]
    {
    \scalebox{0.85}{\usetikzlibrary{shapes,arrows.meta,shadows,positioning}

\tikzset{
  frame/.style={
    rectangle, draw,
    text width=4em, text centered,
    minimum height=4em,fill=white,
    rounded corners=1mm,
  },
  line/.style={
    draw, -{Latex},rounded corners=1mm,
  }
}

\begin{tikzpicture}[font=\small\sffamily\bfseries,very thick,node distance = 4cm]
\node [frame] (agent) {Agent};
\node [frame, right of=agent, node distance=3cm] (action features) {Action\\Features};
\node [frame, below of = action features, node distance=3cm] (environment) {MECH\\ENV};
\draw[->] (agent) -- (action features)
node[midway,above] {$a_t$};
\node [frame, right of =action features, node distance = 3 cm , fill=blue!20] (ctrl dc motor) {CTRL+\\DC Motor};
\draw[->] (action features) -- (ctrl dc motor)
node[midway,above] {$F^\text{desired}_t$};
\draw[->] (ctrl dc motor) -- ++ (0.95,0) |- (environment)
node[left, pos=0.25, align=left] {$F_{t+1}$};;
\node [frame, above of = action features, node distance = 2 cm] (actuated body vel) {Actuated\\Velocity};
\draw[->] (-1.7,0.32) |- (actuated body vel);
\draw[->] (actuated body vel)-| (ctrl dc motor)
node [midway,above] {$w_t$};
\draw[-, dashed] (1,-4) -- (1,-2);

\draw[->] (2.2,-2.5) -- (1,-2.5) node[above, pos=0.5, align = right] {$r_{t+1}$};
\draw[->] (2.2,-3.5) -- (1,-3.5)  node[above, pos=0.5, align = right] {$s_{t+1}$};

\draw[->] (1,-2.5) -- (-1.4,-2.5)    
 -- (-1.4,0) node[midway,right] {$r_{t}$}-- (-0.8,0);

\draw[->] (1,-3.5) -- (-1.7,-3.5) -- (-1.7,0.32) node[midway,left] {$s_{t}$} -- (-0.8,0.32); 
\end{tikzpicture}}
    \label{Fig Extended Reinforcement Learning Block Diagram A}
    }
    \\
    \subfigure[An illustration of the CTRL+DC Motor component in the CO-DRL framework.]
    {
    \scalebox{0.9}{\usetikzlibrary{shapes,arrows.meta,shadows,positioning}

\tikzset{
  frame/.style={
    rectangle, draw,
    text centered,
    minimum height=2em,fill=white,
    rounded corners=1mm,
  },
  line/.style={
    draw, -{Latex},rounded corners=1mm,
  }
}

\tikzstyle{sum} = [draw, circle, node distance=1cm]

\begin{tikzpicture}[font=\small\sffamily\bfseries,very thick,node distance = 4cm]
\coordinate(Fd);
\node[frame, right of=Fd, node distance = 1cm,fill](1/kT) {$\frac{1}{k_T}$};
\node[sum, right of=1/kT, node distance = 1cm](sum) {+};
\draw[->] (Fd) node[midway,above] {$F^\text{desired}_t$}-- (1/kT);
\draw[->] (1/kT) -- node[midway,above]{$i^{\textrm{ref}}_t$} (sum);
\node[frame, right of = sum, node distance = 1.5cm, fill=red!20](ctrl){CTRL};
\draw[->] (sum) -- node[midway,above] {$e_t$} (ctrl);
\node[frame, right of = ctrl, node distance = 2cm](dc motor) {DC Motor};
\draw[->] (ctrl)  -- node[midway,above] {$u_t$} (dc motor);
\coordinate[right of = dc motor, node distance = 0.95cm] (inew);
\draw[-](dc motor) -- (inew);
\node[frame, right of = inew, node distance = 0.8cm](kT){$k_T$};
\draw[->] (inew) -- node[midway, above] {$i_{t+1}$} (kT);
\node[right of = kT, node distance = 1.0cm](Fout){$F_{t+1}$};
\draw[->] (kT) --  (Fout);
 \coordinate (tmp inew) at (4.5,-2);
\draw[-, dashed] (4.5,-1.5)--(4.5,-2.5);
\draw[->] (inew)|- (tmp inew);
\draw[->] (tmp inew) -| (sum) node[left,pos=0.8,align=left] {$i_t$} node[right,pos=0.9,align=right] {$-$};
 \coordinate [above of=dc motor, node distance = 1cm] (w);
 \draw[->] (w) -- node[above,pos=-0.5,align=right] {$w_t$} (dc motor);
  \coordinate [above of=w, node distance = 0.3cm] (tmp w);
  \draw[-] (w) -- (tmp w);
  \draw[->] (w) --+ (0,-0.1) -| (ctrl.north);
\end{tikzpicture}}
    \label{Fig Extended Closed Loop Tracking Control}
    }
    \\
    \subfigure[An illustration of the CTRL component within the CTRL+DC Motor component.]
    {
    \scalebox{0.9}{\usetikzlibrary{shapes,arrows.meta,shadows,positioning}

\tikzset{
  frame/.style={
    rectangle, draw,
    text centered,
    minimum height=2em,fill=white,
    rounded corners=1mm,
  },
  line/.style={
    draw, -{Latex},rounded corners=1mm,
  }
}

\tikzstyle{sum} = [draw, circle, node distance=1cm]

\begin{tikzpicture}[font=\small,very thick,node distance = 4cm]
    \node(e_in){$e_t$};
    \node[frame, right of=e_in, node distance = 2cm](integral){$\frac{\Delta t}{1-z^{-1}}$};
    \node[below of = integral, node distance = 5mm](integral_text){Integrator};
    \draw[->] (e_in) -- (integral);
    \node[frame, below of=integral, node distance = 2cm](derivative){$\frac{1-z^{-1}}{\Delta t}$};
    \node[below of = derivative, node distance = 5mm](derivative_text){Derivative};
    \node[frame, right of = integral, node distance = 2cm](kI){$k_I$};
    \draw[->] -- (integral) -- node[midway, above] {$I_{e_t}$} (kI);
    \node[frame, right of = derivative, node distance = 2cm](kD){$k_D$};
    \draw[->] -- (derivative) -- node[midway, above] {$D_{e_t}$} (kD);
    \node[frame, above of=kI, node distance = 2cm](kP){$k_P$};
    \node[sum, right of=kI, node distance = 2cm](sum){+};
    \draw[->] (kP.east) -- (sum);
    \draw[->] (kI.east) -- (sum);
    \draw[->] (kD.east) -- (sum);
    \coordinate [right of=e_in, node distance = 1cm] (eI);
    \draw[->] (eI) |- (kP);
    \draw[->] (eI) |- (derivative);
    \node[frame, right of=kP, node distance = 2cm](kE){$k_E$};
    \node[above of=kE, node distance = 1.2cm](wt){$w_t$};
    \draw[->] (wt) -- (kE);
    \draw[->] (kE) -- (sum);
    \node[frame, right of=sum, node distance = 1.2cm](sat)
    {$\mathrm{sat}$($\cdot$,$\cdot$,$\cdot$) };
    \draw[->] (sum) -- (sat);
    \node[below of=sat, node distance = 1cm](ubound){(-$u_{\textrm{max}}$,+$u_{\textrm{max}}$)};
    \draw[->] (ubound) -- (sat);
    \node[right of=sat, node distance = 1.5cm](uout){$u_t$};
    \draw[->] (sat) -- (uout);
\end{tikzpicture}}
    \label{Fig Controller Structure}
    }
    \caption{A block diagram of the proposed CO-DRL framework for autonomous systems.}
    \label{Fig Extended Reinforcement Learning Block Diagram} 
\end{figure}
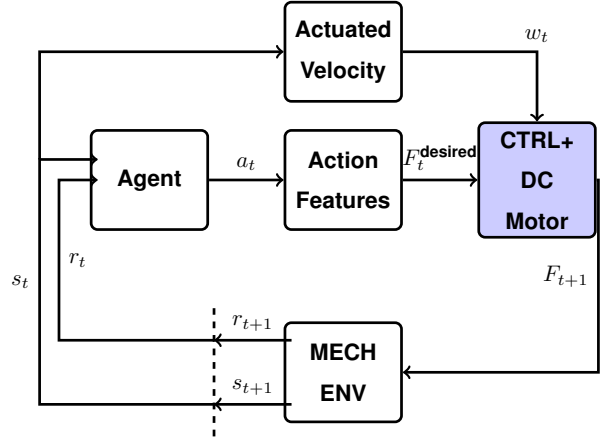
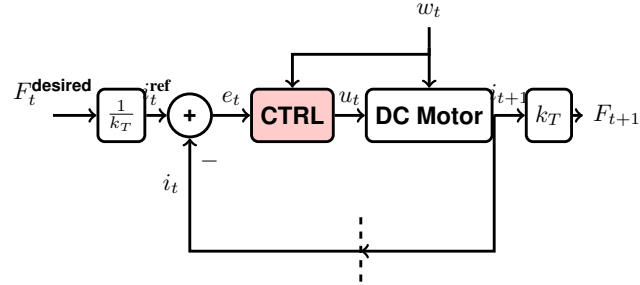
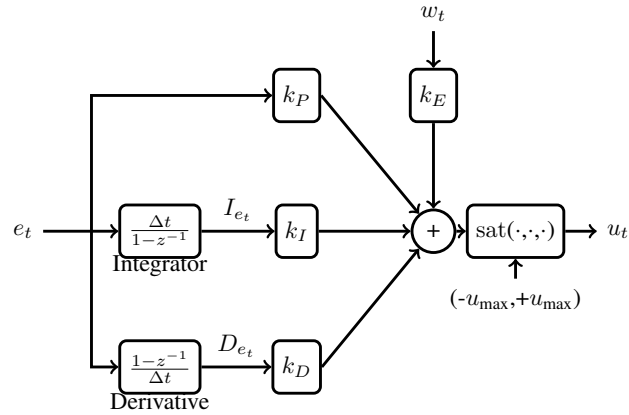

Our proposed approach is illustrated in Fig. \ref{Fig Extended Reinforcement Learning Block Diagram}. Fig. \ref{Fig Extended Reinforcement Learning Block Diagram A} illustrates a novel extension of the DRL framework to address uncertainty in action execution by optimizing system operation through key components: \emph{Action Features}, \emph{Actuated Velocity}, and \emph{CTRL+DC Motor}. Specifically, in addition to feeding the current state $s_t$ and reward $r_t$ into the \emph{Agent} component, $s_t$ is also processed by the \emph{Action Features} component, which extracts the velocity associated with the actuated body via the DC motor \eqref{eq: Actuated Velocity}. This velocity is then fed into the \emph{CTRL+DC Motor} component. The agent’s current action $a_t$ is mapped to a desired force/torque via the \emph{Action Features} component \eqref{eq:Action2desiredForce}, which is subsequently passed to \emph{CTRL+DC Motor}. The \emph{CTRL+DC Motor} component is a key enhancement that determines the actual force/torque applied to the main mechanical environment (\emph{MECH ENV}), which in turn generates the next state $s_{t+1}$ (as per \eqref{eq: state update}) and the corresponding reward $r_{t+1}$ (as per \eqref{eq: reward update}). Further details on the \emph{CTRL+DC Motor} component are provided in Fig. \ref{Fig Extended Closed Loop Tracking Control} and elaborated on in the following discussion.
      
In Fig. \ref{Fig Extended Closed Loop Tracking Control}, we illustrate the implementation of the feedback tracking control loop for tracking the applied force/torque. The DC Motor component models the electrical equation for updating the electrical current \eqref{eq: current update}, while the CTRL component governs the control process based on the error between the reference electrical current $i^\textrm{ref}_t$ and the actual current $i_t$ \eqref{eq: current error}. The actuated velocity $w_t$ is fed into both the DC Motor and CTRL components, though with distinct roles. When fed into the DC Motor component, $w_t$ influences the motor speed (an internal state) and generates an actual BEMF voltage within the electrical circuit. Conversely, when fed into the CTRL component, $w_t$ is measured (or estimated) and incorporated into the control law by multiplying it by the constant $k_E$. This artificially generates the BEMF (based on \eqref{eq: Voltage WRT angle vel}), which is then integrated into the PID control structure \eqref{eq: PID control signal with BEMF and SAT}. Further details on the PID control law structure within the CTRL component are provided in Fig. \ref{Fig Controller Structure} and discussed in the following section.

In Fig. \ref{Fig Controller Structure}, we illustrate the CTRL component and the PID control law structure \eqref{eq: PID control signal with BEMF and SAT}, incorporating the time-shift relation in the z-domain:
\begin{equation}\label{eq: z shift delay}
    z^{-1}\{x_t\}=x_{t-1}.
\end{equation}
We represent the integral using the transfer function:
\begin{equation} \label{eq: z Integral TF BKWD}
    \frac{\Delta t}{1-z^{-1}},
\end{equation}
and the derivative by the transfer function:
\begin{equation} \label{eq: z Derivative TF BKWD}
    \frac{1-z^{-1}}{\Delta t}.
\end{equation}
The proportional, integral and derivative parts are multiplied by the coefficients $k_P$, $k_I$, and $k_D$, respectively, and summed with the the term $k_E w_t$. The final control signal $u_t$ is then passed through the Sat component to ensure it remains within predefined bounds. 

\subsubsection{Pseudocode of the CO-DRL 
Algorithm}

The CO-DRL algorithm is summarized in the following pseudocode in Algorithm \ref{alg: pseudocode1}. Let $s_0$ represent the initial state of the mechanical and original environment, and $i_0$ the initial electrical current supplied to the DC motor. The algorithm starts at discrete time $t=0$ and loops until termination, which is indicated by the Boolean variable \emph{terminated}. Termination is determined by the environment's conditions, such as maximum time iterations or the success/failure of meeting specific state and action criteria. These conditions are defined by the Gym library for each environment (prior to our development). At each step, an action $a_t$ is chosen according to the DRL algorithm designed for the environment. This action is then converted into a desired force/torque $F^\text{desired}_t$ through the ActionFeature. While the ActionFeature is implemented within each Gym environment, we refer to it as the desired force/torque, rather than the actual force/torque. The desired force/torque is converted into a reference electrical current $i^{\text{ref}}_t$, and the error $e_t = i^{\text{ref}}_t - i_t$ is computed. From the current state $s_t$, the velocity of the actuated body $w_t$, which is related to the DC motor speed, is obtained through the ActuatedVelocity function. The actual force/torque $F_{t+1}$ is generated based on the error $e_t$ and the velocity $w_t$ by the PID controller and the DC motor electrical equations, as described in the DCMotorENVStep$(e_t, w_t)$ function below the main loop (our new development for each environment). Using the actual force/torque $F_{t+1}$, we update the state $s_{t+1}$, reward $r_{t+1}$, observations $o_{t+1}$, time step, and the termination flag \emph{terminated}.        

\begin{algorithm}[H]
\small
\label{alg: pseudocode1}
\SetAlgoLined
 Initialization $s_0$, $i_0$ \;
 t=0\;
 \While{Not terminated}{
  Choose action: $a_t$\;
  Map action to desired force: $F^\text{desired}_t = \texttt{ActionFeature}(a_t)$\;
  Set reference electrical current: $i^{\text{ref}}_t = F^\text{desired}_t/k_T$\;
  Compute error: $e_t = i^{\text{ref}}_t-i_t$\;
  Extract actuated body velocity: $w_t = \texttt{ActuatedVelocity}(s_t)$\;
  Get actual force: $F_{t+1} = \texttt{DCMotorENVStep}(e_t, w_t)$\;
  Update state: $s_{t+1}=\texttt{EnvStateUpdate}(s_t,F_{t+1})$\;
  Get new reward: $r_{t+1} = \texttt{Reward}(s_{t+1},s_t, F_{t+1})$ \;
  Get new observations: $o_{t+1} = \texttt{getObs}(s_{t+1},s_t, F_{t+1})$\;
  $t=t+1$\;
  Update \textit{terminated} flag as necessary\;
 }

\SetAlgoLined
  \SetKwFunction{FMain}{DCMotorENVStep}
  \SetKwProg{Fn}{Function}{:}{}
  \Fn{\FMain{$e_t$, $w_t$}}{
        Get previous error $e_{t-1}$, and previous error integral $I_{e_{t-1}}$ saved in \texttt{DCMotorENV}\;
        Update integral: $I_{e_t}=I_{e_{t-1}}+e_{t}{\Delta t}$\;
        Update derivative: $D_{e_t}=\frac{e_t-e_{t-1}}{\Delta t}$ \;
        Compute voltage input: \newline $u_t=\text{sat}(k_P e_t + k_I I_{e_t} + k_D D_{e_t} + k_E w_t,-u_\text{max},u_\text{max})$\;
        Update current: $i_{t+1} = i_t + \frac{u_t-V^{BEMF}_t-Ri_t}{L} {\Delta t}$\;
        Compute actual force: $F_{t+1}=k_T i_{t+1}$\;
        \KwRet $F_{t+1}$\;
  }
 \caption{The CO-DRL Algorithm}
\end{algorithm}
 
\vspace{0.3cm}
\subsubsection{Open Source Software}
\label{ssec:open_source}

For the benefit of researchers and developers in related fields, we developed an open source software implementation of the CO-DRL framework for autonomous systems. Practitioners in related fields are welcome to integrate our implementation in their working environment. Our implementation was developed using Python and is available at GitHub (see link in \cite{web:OrenGit}). The implementation details of the open source software are described in the Appendix.

\section{Experimental Results}
\label{sec:experiemnts}

In this section, we present a comprehensive experimental study to evaluate the performance of the CO-DRL algorithm, which accounts for uncertainties in action execution due to the internal dynamics of AI-driven autonomous systems. We conduct simulations using various fixed values for the PID controller's coefficients, representing a manual tuning approach. These simulations reflect scenarios where the PID controller within the DC motor functions as a black box from the DRL designer's perspective.

\begin{figure}[H]
    \centering
    \subfigure[Pendulum environment]
    {
        \includegraphics[width=2in,height=1.5in]{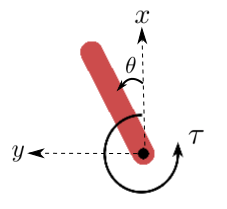}
        \label{Fig illustration of the Pendulum environment}
    }
    \subfigure[Mountain Car environment]
    {
        \includegraphics[width=2in,height=1.5in]{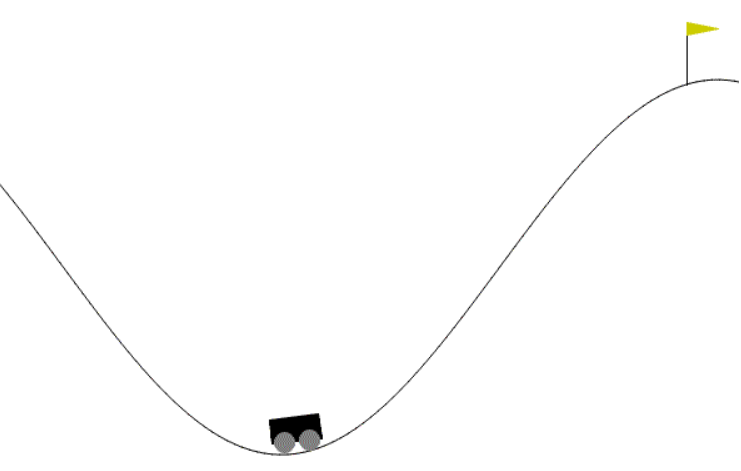}
        \label{Fig illustration of the Mountain Car environment}
    }
    \\
    \subfigure[Acrobot environment]
    {
        \includegraphics[width=2in,height=1.5in]{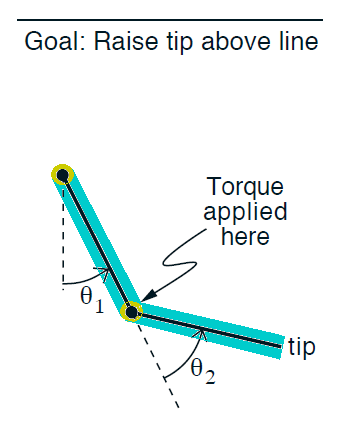}
        \label{Fig illustration of the Acrobot environment}
    }
    \subfigure[Cartpole environment]
    {
        \includegraphics[width=2in,height=1.5in]{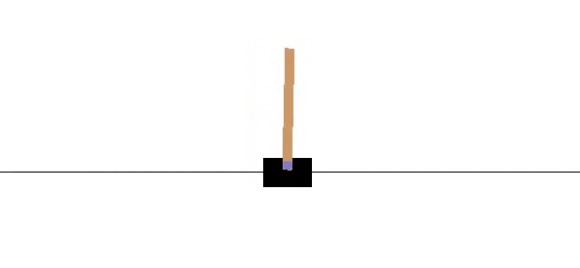}
        \label{Fig illustration of the Cartpole environment}
    }    
    \caption{Illustrations of the OpenAI Gym environments that have been restructured and augmented in this work to capture the complexities of real-world action uncertainties.}
    \label{Fig Illustrations of gym environments}
\end{figure}

Table \ref{tab: Gym ENV Action Alg} provides a summary of all Gym environments used in our implementation, and Fig. \ref{Fig Illustrations of gym environments} provides the illustrations of those Gym environments. It details the following aspects: the type of desired action (continuous or discrete), the DRL algorithm employed to determine the agent's actions (e.g., DDPG, DQN, etc.), and the Python platform used for implementing the DRL algorithm (e.g., TensorFlow, PyTorch). 

The parameters of the DC motor's electrical circuit remain consistent across all environments with the following values: resistance $R=1$ $\Omega$, inductor $L=0.1$ H, torque constant $k_T=1$ Volt per meter per second (or radian per second).

\begin{table}[H]
    \centering
     \caption{Summary of Implemented Gym's Environments}
    \begin{tabular}{||c|c|c|c||}
       \hline
        \thead{Gym ENV}  & \thead{Disc./Cont.\\Action} & \thead{Algorithm} & \thead{Platform}\\ 
       \hline
        Pendulum  & Continuous & DDPG & TesnsorFlow\\
       \hline
        \makecell{Mountain Car\\(Disc.)}  & Discrete & \makecell{Ep. Semi-Grad\\SARSA\\+Tile Coding} & \makecell{tile3 \\\cite{sutton1995generalization,Bogdanski_Episodic_SemiGrad_SARSA}}\\
       \hline
        \makecell{Mountain Car\\(Cont.)}  & Continuous & DDPG & Pytorch\\
       \hline
        Acrobot  & Discrete & DQN & Pytorch\\
        \hline
        Cartpole  & Discrete & PPO & TesnsorFlow\\
       \hline
    \end{tabular}
    \label{tab: Gym ENV Action Alg}
\end{table}

\subsection{Solving the Pendulum Environment}

In the Pendulum environment, a pendulum is attached to a fixed point at one end, while the other end is free. which is allowed to move along the x-axis, as illustrated in Fig. \ref{Fig illustration of the Pendulum environment}.

The goal of this environment is to apply torque to the free end of the pendulum to swing it upright, positioning its center of gravity above the fixed point. The state space consists of the pendulum’s angle \(\theta \in [-\pi, \pi]\) (radians) and angular velocity \(\dot{\theta} \in [-8,8]\) (rad/sec). The observation space represents the pendulum’s position in Cartesian coordinates, where \(x = \cos(\theta)\) and \(y = \sin(\theta)\), with \(x, y \in [-1,1]\). The action space is a continuous torque input ranging from \([-2,2]\) Nm. In our approach, action feature extraction is defined by: 
\begin{equation}
    F^\text{desired}_t=\text{sat}(a_t,-\texttt{max\_torque},+\texttt{max\_torque}),
\end{equation}
where $a_t$ is the action input, $\texttt{max\_torque}$=2 denotes the nominal value of the maximal desired torque to be applied, and $F^\text{desired}_t$ the result of the desired torque to be fed into the PID control to generate the actual torque $F_{t+1}$.
The reward is defined by:
\begin{equation}
    r_{t+1}=-\theta_{t}^2-0.1 \dot {\theta_t}^2-0.001F_{t+1}^2.
\end{equation}
In Gym, the reward is updated before the state update. The minimum reward is -16.27, and the maximum is zero (when the pendulum is upright, with zero velocity and no torque). The episode is truncated after 200 time steps, each lasting 0.05 seconds.

The agent’s desired action architecture is DDPG. The actor's neural network (NN) consists of: an input layer (2 units), two hidden layers (64 units each, with Rectified Linear Unit (ReLU) activation), and an output layer (1 unit, with Tanh activation, multiplied by 2). The critic's NN has the following structure: State input layer (2 units), state output layer (1 unit, ReLU), action input layer (1 unit), action output layer (32 units, ReLU), followed by concatenated state and action layers, two hidden layers (256 units each, ReLU), and a final output layer (1 unit). Other parameters include: learning rates \(\alpha_\text{actor} = 0.001\) and \(\alpha_\text{critic} = 0.002\), discount factor \(\gamma = 0.99\), target NN update parameter \(\tau = 0.005\), 500 episodes, and the ADAM optimizer.

\begin{figure}[htbp]
    \centering
    \subfigure[Average Episodic Reward]
    {
        \includegraphics[width=3in,height=1.3in]{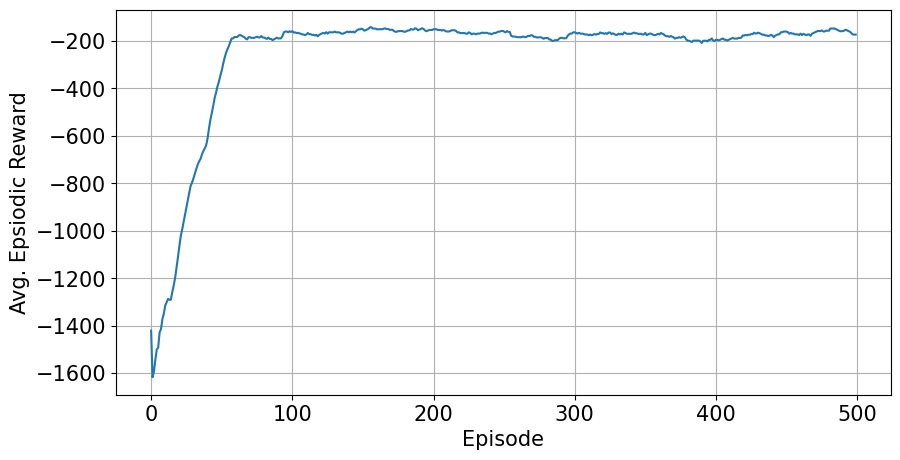}
        \label{Fig Pendulum Reward Kp1 Ki20 Kd1e-6}
    }
    \subfigure[Desired Torque and Actual Torque]
    {
        \includegraphics[width=3in,height=1.3in]{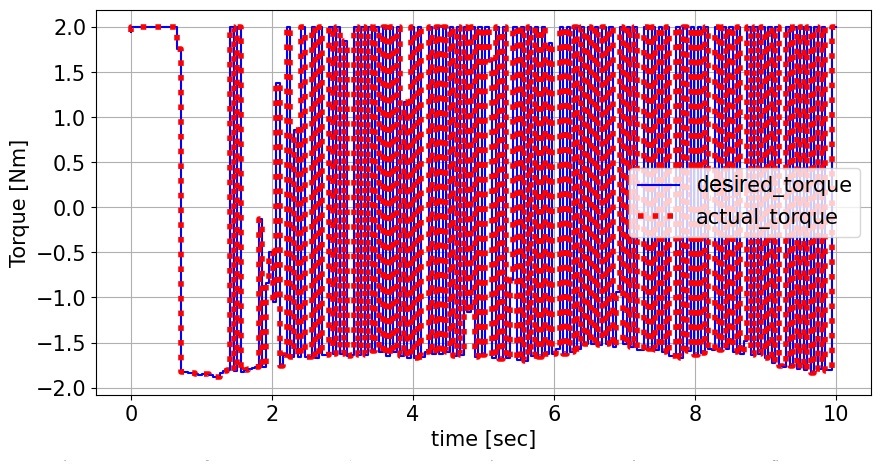}
        \label{Fig Pendulum Action Kp1 Ki20 Kd1e-6}
    }
    \subfigure[Vertical Position $x=\cos(\theta)$ (Positive Direction: Up)]
    {
        \includegraphics[width=3in,height=1.2in]{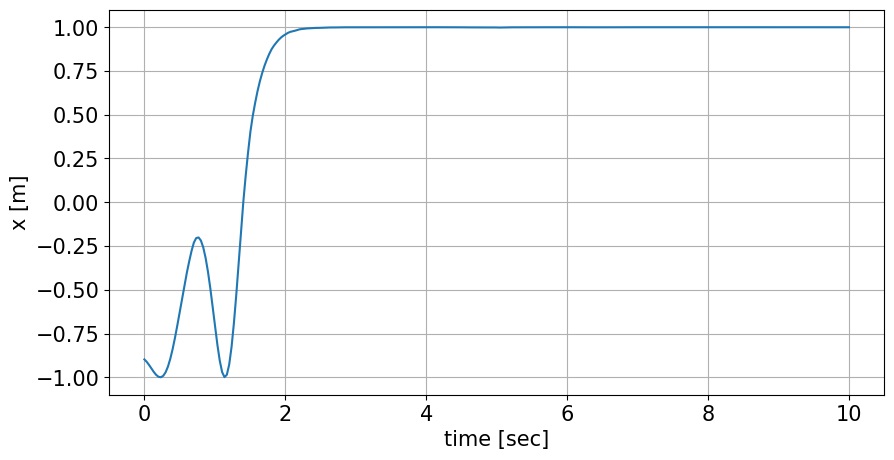}
        \label{Fig Pendulum x Kp1 Ki20 Kd1e-6}   
    }
    \subfigure[Horizontal Position $y=\sin(\theta)$ (Positive Direction: Left)]
    {
        \includegraphics[width=3in,height=1.2in]{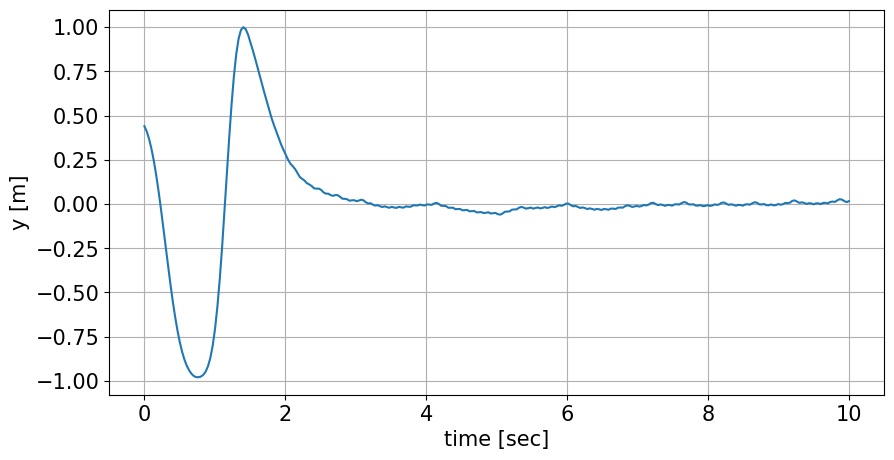}
        \label{Fig Pendulum y Kp1 Ki20 Kd1e-6}   
    }
    \subfigure[Angle $\theta=\textrm{atan2}(y,x)$]
    {
        \includegraphics[width=3in,height=1.2in]{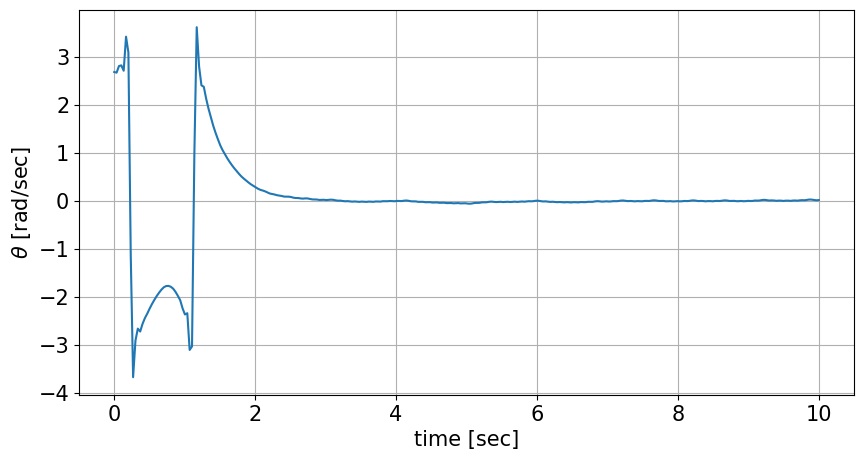}
        \label{Fig Pendulum Theta Kp1 Ki20 Kd1e-6}
    }
    \subfigure[Angular Velocity $\dot{\theta}$]
    {
        \includegraphics[width=3in,height=1.2in]{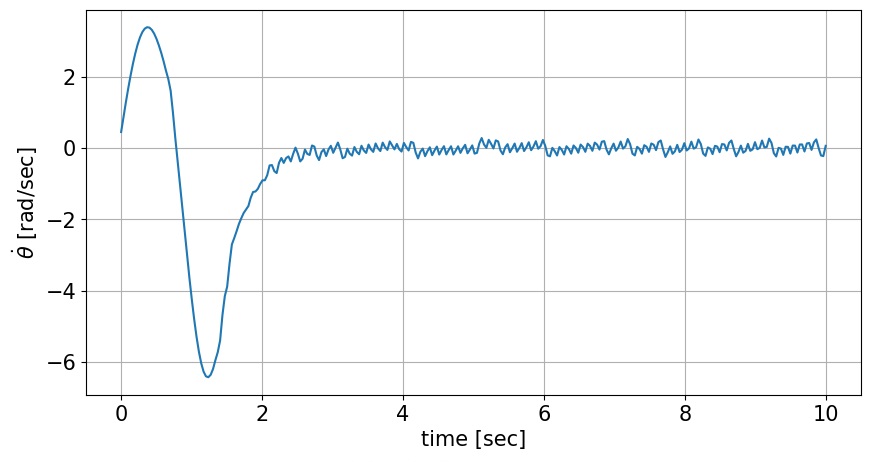}
        \label{Fig Pendulum Theta dot Kp1 Ki20 Kd1e-6} 
    }
    \caption{Simulation results for the Pendulum environment. Simulation parameters: $K_P=$1, $K_I=$20, $K_D=$1e-6.}
    \label{Fig Pendulum Results Kp1 Ki20 Kd1e-6}
\end{figure}

The results are shown in Fig. \ref{Fig Pendulum Results Kp1 Ki20 Kd1e-6}. With PID coefficients \(K_P = 1\), \(K_I = 20\), and \(K_D = 1e-6\), the desired torque was tracked almost perfectly (Fig. \ref{Fig Pendulum Action Kp1 Ki20 Kd1e-6}), yielding similar agent performance, reward (Fig. \ref{Fig Pendulum Reward Kp1 Ki20 Kd1e-6}), and pendulum observations (Figs. \ref{Fig Pendulum x Kp1 Ki20 Kd1e-6} to \ref{Fig Pendulum Theta dot Kp1 Ki20 Kd1e-6}) compared to the ideal environment. Within 10 seconds, the agent successfully held the pendulum vertically (\(x = 1\), \(y, \theta, \dot{\theta} = 0\)).

\subsection{Solving the Mountain Car Environment}

This subsection covers solutions for both versions of the Mountain Car environment (Fig. \ref{Fig illustration of the Mountain Car environment}): the discrete version (Mountain Car in Gym) and the continuous version (Mountain Car Continuous in Gym).

The goal is to drive the car to the top of the right hill, starting from the bottom of the left hill. The state (and observation) space is continuous, including the car's position along the x-axis \( x \in [-1.2, 0.6] \) (in meters) and velocity \( v \in [-0.07, 0.07] \) (in m/sec). The action space applies a force (in N) between \(-1\) and \(+1\). In the continuous version, the action space is \( [-1,1] \), defining the action feature as \( F^\text{desired}_t = \text{sat}(a_t, -1, +1) \). In the discrete version, there are three action indices \( a_t \in \{0,1,2\} \), mapping to forces \( F^\text{desired}_t = a_t - 1 \in \{-1,0,1\} \). State updates in Gym's Mountain Car environment follow the dynamic equations below \cite{SuttonBarto2018}:
\begin{align} 
    v_{t+1} &= v_t+ C_f F_{t+1} - C_g \cos(3x_t), \label{eq: velocity update mountain car}\\
    x_{t+1} &= x_t+v_{t+1}. \label{eq: position update mountain car} 
\end{align}
The new velocity \( v_{t+1} \) and position \( x_{t+1} \) are clipped to \( [-0.07,0.07] \) and \( [-1.2, 0.6] \), respectively. The actual applied force is \( F_{t+1} \) (\( F_{t+1} = F^\text{desired}_t \) in the original environment). Constants include \( C_f \) for force (\texttt{self.force=0.001} in the discrete version, \texttt{self.power=0.0015} in the continuous version) and \( C_g \) for gravity (\texttt{self.gravity=0.0025}). Since position updates \eqref{eq: position update mountain car} mix terms with different units—position in [m] and velocity in [m/sec]—Gym assumes a 1-second integration time, which may be too coarse for state updates and PID control. To refine this, we adopt a continuous formulation using integral equations:
\begin{align} 
    v(t+\Delta t) &= v(t)+\int_{t}^{t+\Delta t}\left [C_f F(t') - C_g \cos(3x(t'))\right ]dt', \label{eq: velocity update mountain car integral}\\
    x(t+\Delta t) &= x(t)+\int_{t}^{t+\Delta t}v(t'+\Delta t)dt'. 
    \label{eq: position update mountain car integral} 
\end{align}
Integration occurs over the interval \( [t, t+\Delta t] \) with \( \Delta t=1 \) second (as in the original environment), but the integrals are approximated using a Riemann sum with a finer partition \( \Delta t' < 1 \) (e.g., 0.05). This approach, detailed in Algorithm \ref{alg: pseudocode2}, introduces an inner loop. The state \( s_t \) updates every \( \Delta t \), while integration occurs at intervals of \( \Delta t' = \Delta t / N \) for some \( N \in \mathbb{N} \). Before entering the loop, the desired force \( F^\text{desired}_t \) is set as a reference, and the current state is stored in a temporary variable \( s' = s_t \). The loop iterates from 0 to \( \Delta t \) (excluding \( \Delta t \)) with step \( \Delta t' \) (similar to \texttt{numpy.arange} in Python). Within each iteration, the temporary velocity \( w' \), actual force \( F' \), and next state \( s' \) are updated. The \texttt{GetActualForce} function encapsulates force computation, including conversion to reference current, error calculation, PID control, and current-to-force conversion. To keep the algorithm concise, step-by-step details are minimized (see Algorithm \ref{alg: pseudocode2} for further explanation).

\begin{algorithm}[h]
\label{alg: pseudocode2}
\SetAlgoLined
 Initialization $s_0$, $\Delta t$\;
 $\Delta t'=\Delta t/N$\;
 t=0\;
 \While{Not terminated}{
    Choose action: $a_t$\;
    $F^\text{desired}_t = \texttt{ActionFeature}(a_t)$\;
    $s'=s_t$\;
    \For{$t'\in$ \texttt{arrange}(start=0, stop=$\Delta t$, step=$\Delta t'$)}{
        $w' = \texttt{ActuatedVelocity}(s')$\;
        $F' = \texttt{GetActualForce}(F^\text{desired}_t, w')$\;
        $s'=\texttt{EnvStateUpdate}(s',F')$\;
        \tcc{e.g., s'+=func(s',F')$\Delta$ t'}\
    }
    $s_{t+1}=s'$\;
    $F_{t+1}=F'$\;
    $r_{t+1} = \texttt{Reward}(s_{t+1},s_t, F_{t+1})$ \;
    $o_{t+1} = \texttt{getObs}(s_{t+1},s_t, F_{t+1})$\;
    $t=t+1$\;
  Update \textit{terminated} flag as necessary\;
 }

 \caption{Inner Loop Integration for State Update}
\end{algorithm}

In the discrete Mountain Car environment, the reward is \(-1\) per time step (\(\Delta t=1\) second) with no additional reward for reaching the goal. In the continuous version, the reward is a negative quadratic function of the applied force: \( r_{t+1} = -0.1 F_{t+1}^2 \), where the \(-0.1\) factor follows Gym's original design for consistency. Additionally, a reward of \( +100 \) is given if the cart reaches the goal. The reward function is defined as:
\begin{equation}
   r_{t+1}=100\cdot [x_{t+1} \geq x_{\textrm{Goal}}]-0.1F_{t+1}^2, 
\end{equation}
where \( x_{\textrm{Goal}}=0.45 \) is the goal position at the top of the right hill, and \( [x_{t+1} \geq x_{\textrm{Goal}}] \) is a Boolean indicator (1 if true, 0 otherwise).

The episode ends if the car's position exceeds the goal ($x_{\textrm{Goal}}=0.45$ in the continuous version, $x_{\textrm{Goal}}=0.5$ in the discrete) or if the episode reaches 999 steps (continuous) or 200 steps (discrete).

For the discrete case, the agent's desired action is modeled using Tile Coding and Episodic Semi-Gradient SARSA. Tile Coding parameters include 8 tilings, 3 actions, 2 state variables, and a total of 1,944 tiles, with a learning rate of 0.0375. The Episodic Semi-Gradient SARSA parameters are a discount factor $\gamma=1.00$, an epsilon-greedy parameter $\varepsilon=0.001$, and 600 episodes.

For the continuous case, the agent's desired action is modeled using DDPG. The actor network consists of an input layer of size 2 (number of observations), followed by two fully connected hidden layers with 64 units each and ReLU activation, and an output layer with one unit and Tanh activation. The critic network processes the state through an input layer of size 2 and a fully connected output layer with 64 units and ReLU activation. The action is processed through a separate input layer of size 1 before being concatenated with the state output. This is followed by a fully connected hidden layer with 64 units and ReLU activation, and a final output layer with one unit. Other parameters include a learning rate of 1e-3 for the actor and 4e-5 for the critic, a discount factor $\gamma=0.85$, target network update parameter $\tau=0.45$, 200 episodes, and the ADAM optimizer.

\begin{figure}[H]
    \centering
    \subfigure[Accumulated Episodic Reward]
    {
        \includegraphics[width=3in,height=1.7in]{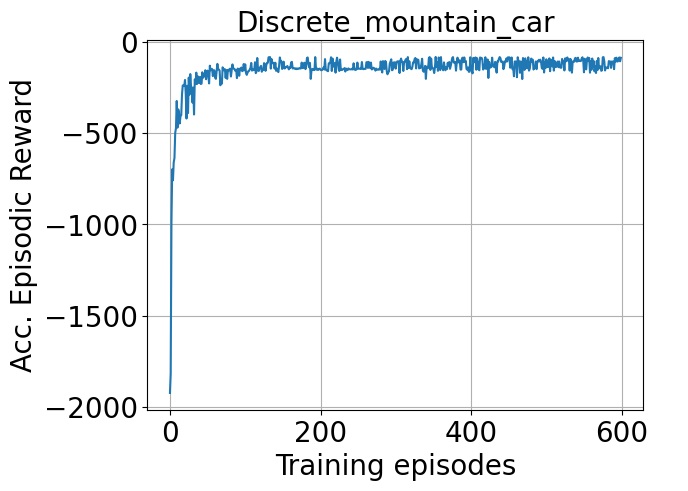}
        \label{Fig Disc Mont Car Reward Kp1 Ki1 Kd1e-2}
    }
    \subfigure[Desired Force and Actual Force]
    {
        \includegraphics[width=3in,height=1.7in]{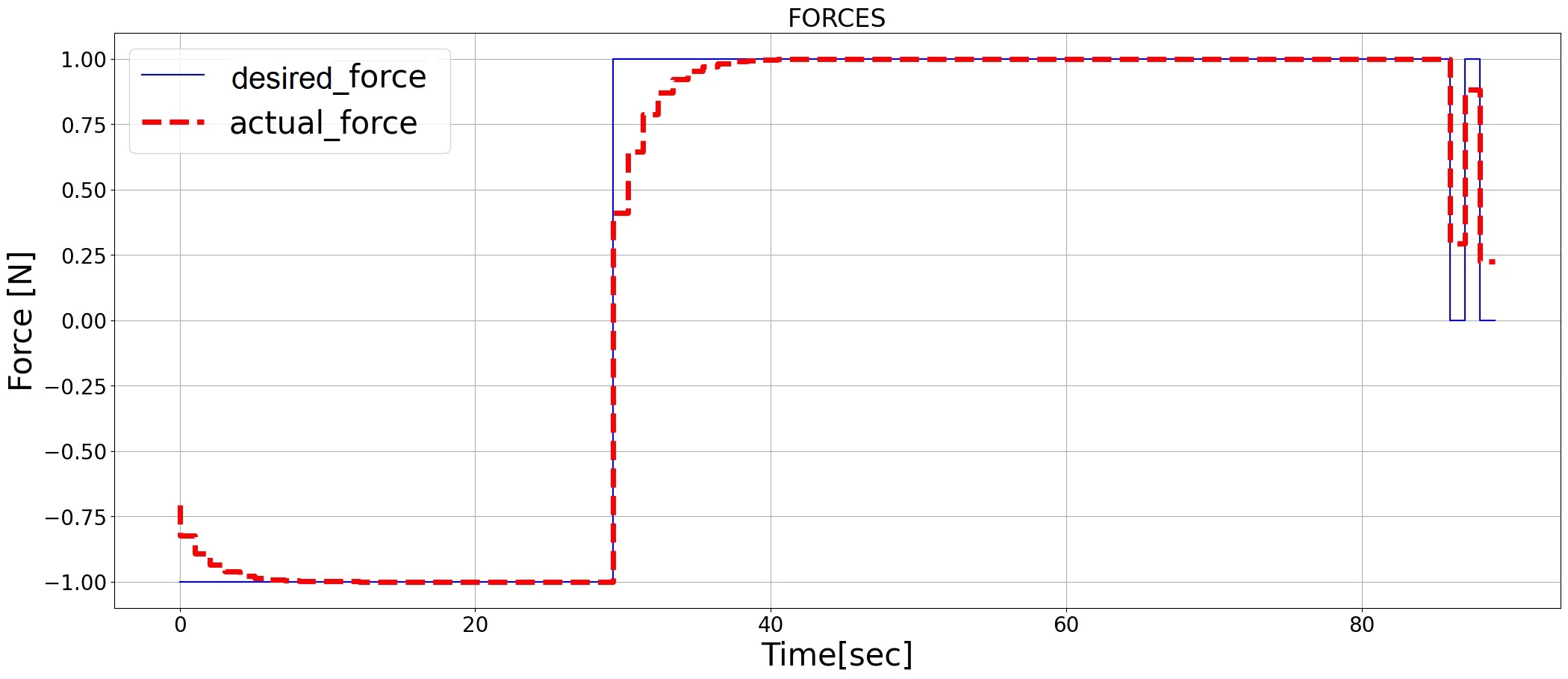}
        \label{Fig Disc Mont Car Action Kp1 Ki1 Kd1e-2}
    }
    \subfigure[Horizontal Position $x$]
    {
        \includegraphics[width=2.5in,height=1.7in]{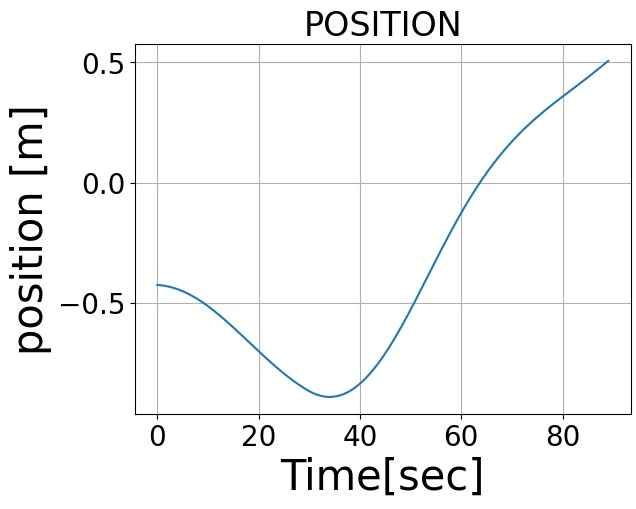}
        \label{Fig Disc Mont Car x Kp1 Ki1 Kd1e-2}
    }
    \caption{Simulation results for the Discrete Mountain Car environment. Simulation parameters: $K_P=$1, $K_I=$1, $K_D=$1e-2.}
    \label{Fig Disc Mont Car Results Kp1 Ki1 Kd1e-2}
\end{figure}

For the discrete version (Fig. \ref{Fig Disc Mont Car Results Kp1 Ki1 Kd1e-2}), results with $K_P=1$, $K_I=1$, and $K_D=1e-2$ show that while force tracking (Fig. \ref{Fig Disc Mont Car Action Kp1 Ki1 Kd1e-2}) was imperfect and took time to converge, the car reached the goal in under 85 seconds (Fig. \ref{Fig Disc Mont Car x Kp1 Ki1 Kd1e-2}), outperforming both the ideal environment (160 sec) and a setup with perfect force tracking (100 sec for $K_P=2$, $K_I=10$, $K_D=1e-6$). The average reward over the last 50 episodes was approximately -114 (Fig. \ref{Fig Disc Mont Car Reward Kp1 Ki1 Kd1e-2}).

\begin{figure}[H]
    \centering
    \subfigure[Accumulated Episodic Reward]
    {
        \includegraphics[width=3.8in,height=4.5in]{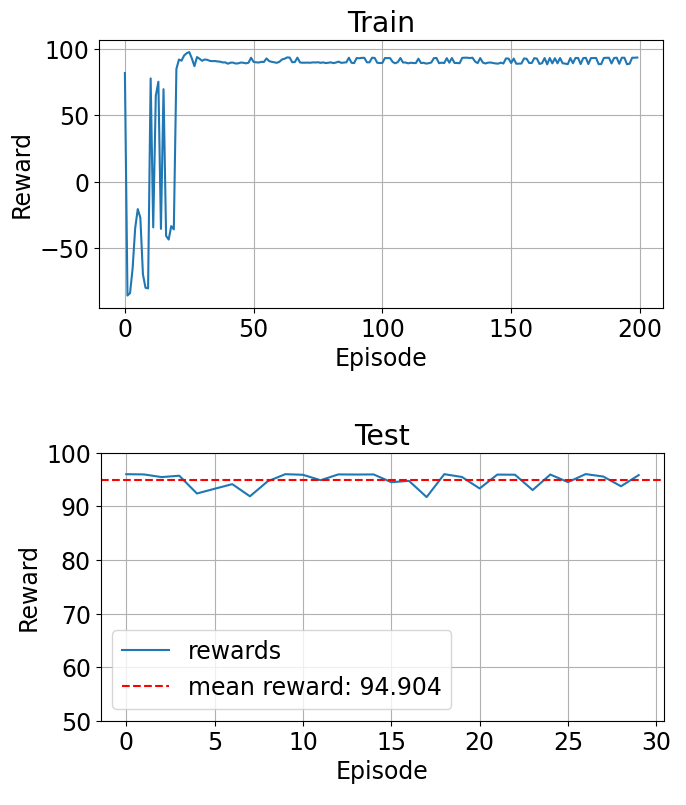}
        \label{Fig Cont Mont Car Reward Kp1e-1 Ki10 Kd1e-3}
    }
    \subfigure[Desired Force and Actual Force]
    {
        \includegraphics[width=3.8in,height=2.2in]{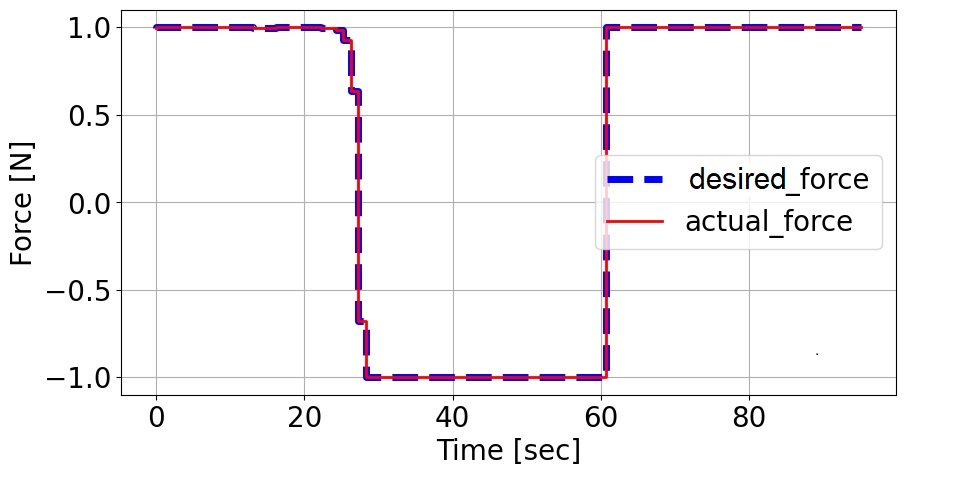}
        \label{Fig Cont Mont Car action Kp1 Ki10 Kd1e-3}
    }
    \subfigure[Horizontal Position $x$]
    {
        \includegraphics[width=3.5in,height=2.3in]{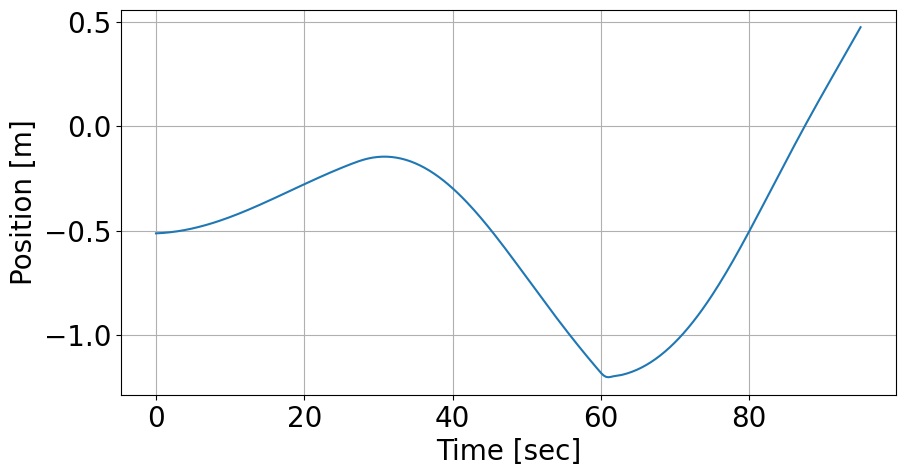}
        \label{Fig Cont Mont Car x Kp1e-1 Ki10 Kd1e-3}
    }
    \caption{Simulation results for the Continuous Mountain Car environment. Simulation parameters: $K_P=$1e-1, $K_I=$10, $K_D=$1e-3.}
    \label{Fig Cont Mont Car Results Kp1e-1 Ki10 Kd1e-3}
\end{figure}

For the continuous version (Fig. \ref{Fig Cont Mont Car Results Kp1e-1 Ki10 Kd1e-3}, using $K_P=1e-1$, $K_I=10$, and $K_D=1e-3$, force tracking was nearly perfect (Fig. \ref{Fig Cont Mont Car action Kp1 Ki10 Kd1e-3}), reward converged to $\sim 95$ (Fig. \ref{Fig Cont Mont Car Reward Kp1e-1 Ki10 Kd1e-3}), and the car reached the goal in 75 seconds (Fig. \ref{Fig Cont Mont Car x Kp1e-1 Ki10 Kd1e-3}).

\subsection{Solving the Acrobot Environment}

Acrobot Environment (Fig. \ref{Fig illustration of the Acrobot environment}) is a mechanical system consists of two connected links forming a chain with one end fixed. An external torque (in [Nm]) actuates the joint between the links, aiming to swing the free end upright.

The State Space includes the first joint's angle $\theta_1$ and angular velocity $\dot{\theta_1}$, along with the relative angle $\theta_2$ and its velocity $\dot{\theta_2}$. The Observation Space represents angles using $\cos{\theta_1}$, $\sin{\theta_1}$, $\cos{\theta_2}$, and $\sin{\theta_2}$.

The action space is discrete with three options, $a_t \in \{0,1,2\}$, defining the desired torque as $F^\text{desired}_t = a_t - 1 \in \{-1,0,1\}$. This torque is processed by a PID controller and DC motor to generate a continuous actual torque $F_{t+1}$. The state updates follow a 4th-order Runge-Kutta method with a 0.2-second interval. We implement a 10-step integration, as in Algorithm \ref{alg: pseudocode2}, similar to the Mountain Car Environment, using a step size of 0.02 seconds.

The agent's desired action is determined using a DQN with a NN comprising an input layer of size 6 (equal to the number of observations), two fully connected hidden layers with 64 and 128 units respectively, both using ReLU activation, and an output layer with 3 units (equal to the number of actions). The model parameters include a learning rate of $\alpha=1e-3$, a discount factor of $\gamma=0.99$, 500 training episodes, and the ADAM optimizer.

\begin{figure}[h]
    \centering
    \subfigure[Average Episodic Reward]
    {
        \includegraphics[width=3in,height=1.7in]{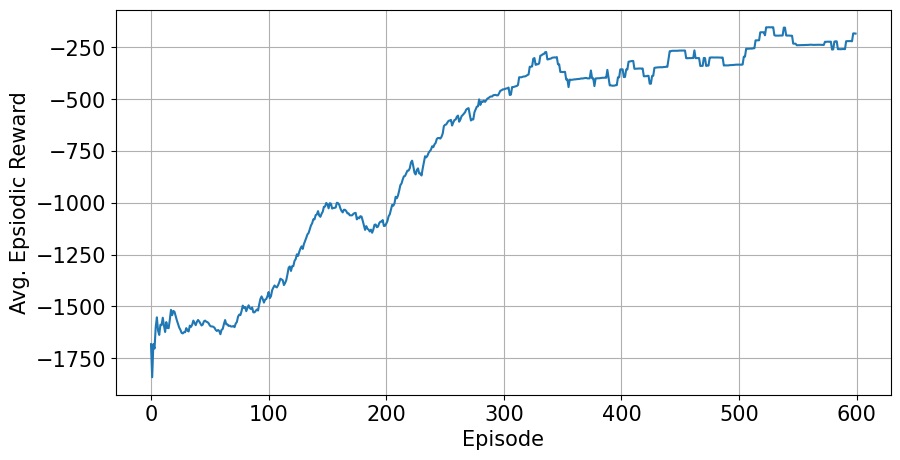}
        \label{Fig Acrobot Reward Kp1 Ki10 Kd1e-6}
    }
    \subfigure[Desired Torque and Actual Torque]
    {
        \includegraphics[width=3in,height=1.7in]{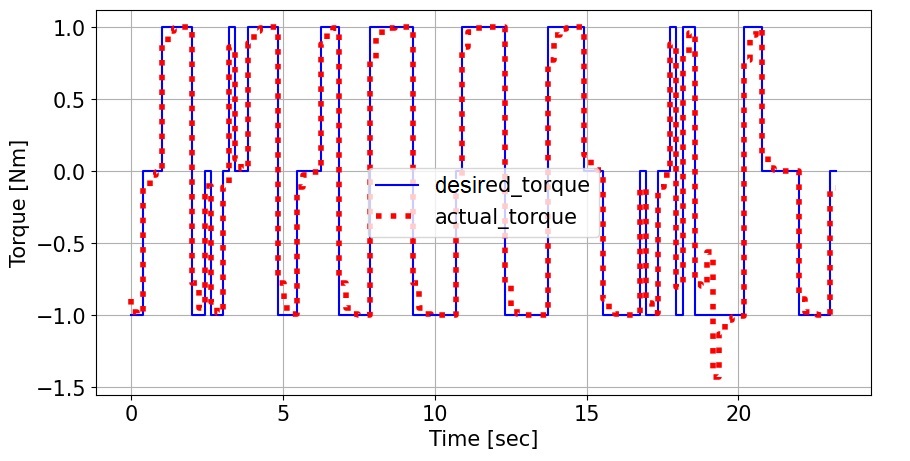}
        \label{Fig Acrobot Action Kp1 Ki10 Kd1e-6}
    }
    \subfigure[$-\cos(\theta_1)-\cos(\theta_1+\theta_2)$]
    {
        \includegraphics[width=2.9in,height=1.7in]{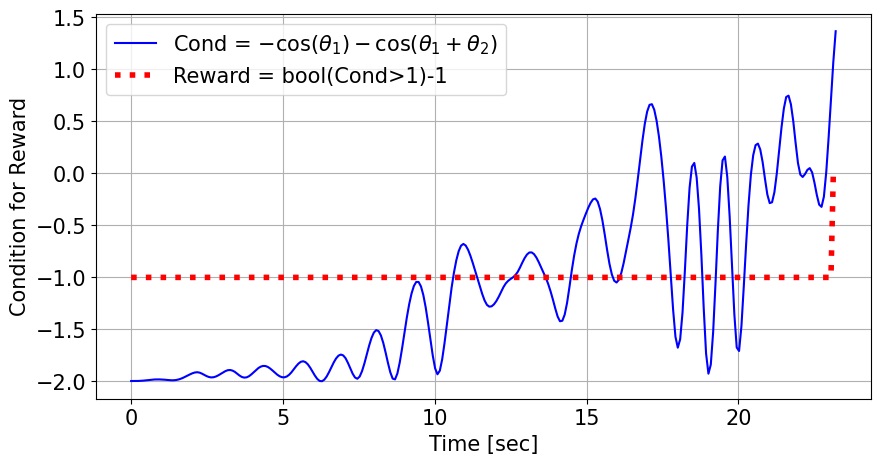}
        \label{Fig Acrobot Cond Kp1 Ki10 Kd1e-6}
    }
    \caption{Simulation results for the Acrobot environment. Simulation parameters: $K_P=$1, $K_I=$10, $K_D=$1e-6.}
    \label{Fig Acrobot Results Kp1 Ki10 Kd1e-6}
\end{figure}

Results are shown in Fig. \ref{Fig Acrobot Results Kp1 Ki10 Kd1e-6}. For $K_P=1$, $K_I=10$, and $K_D=1e-6$, the desired torque is well-tracked (Fig. \ref{Fig Acrobot Action Kp1 Ki10 Kd1e-6}), the reward converges to approximately -200 (Fig. \ref{Fig Acrobot Reward Kp1 Ki10 Kd1e-6}), and the Acrobot reaches the goal in about 25 seconds (Fig. \ref{Fig Acrobot Cond Kp1 Ki10 Kd1e-6}).

\subsection{Solving the Cartpole Environment}
In the Cartpole environment (Fig. \ref{Fig illustration of the Cartpole environment}), a pole is attached to a cart that moves along the x-axis. The state space includes the cart’s position \( x \) and velocity \( \dot{x} \), as well as the pole’s angle \( \theta \) and angular velocity \( \dot{\theta} \). At each time step, the agent applies a force of either -10 N (left) or +10 N (right):
\begin{equation}
    F^\text{desired}_t=\
    \begin{cases}
        -10, & a_t=0,\\
        +10, & a_t=1.
    \end{cases}
\end{equation}
Although the action space is discrete, mapping to desired forces of \(\pm 10\) N, the actual force \( F_{t+1} \) from the DC motor is continuous, regulated by a PID controller. The pendulum starts upright, and the objective is to keep it balanced. The agent earns a reward of \( +1 \) for each time step the pole remains upright. The cart's position is constrained to \([-2.4, 2.4]\) meters, and the pole's angle to \([-0.2095, 0.2095]\) radians (or \([-12^\circ, 12^\circ]\)); exceeding these limits results in failure. The reward function is given by:
\small
\begin{equation}
    r_{t+1}=
    \begin{cases}
        1, & x_{t+1}\in [-2.4,2.4] \text{ and } \theta_{t+1}\in [-0.2095, 0.2095] \\
        0, & x_{t+1}\notin [-2.4,2.4] \text{ or } \theta_{t+1}\notin [-0.2095, 0.2095].
    \end{cases}
\end{equation}
\normalsize
The episode ends when the pole falls ($r_{t+1}=0$) or remains upright for $N$ time steps—500 in cartpole-v1, 200 in cartpole-v0—each lasting 0.02 seconds. The maximum reward per episode is $N$.

To evaluate our algorithm, we modified the cartpole environment and PID controller. First, we adjusted the arm length \( r \) in the torque-force and velocity relations to keep the desired action signal within a reasonable range. Instead of tracking a torque of \( \pm 10 \) Nm (with \( r = 1 \) m), we set \( r = 0.15 \) m, resulting in a lower torque of \( \pm 1.5 \) Nm. Second, we reshaped the reward function by penalizing the agent (reducing the reward from 1 to 0) whenever the cart's position exceeded \(|x| > 0.1\), without terminating the episode. This encourages the agent to keep the cart near \( x=0 \) while avoiding the termination zone at \(|x| > 2.4\). Lastly, we introduced a feedforward term \( K_\textrm{ff} i^\textrm{ref}_t \) into the control signal \( u_t \) to improve tracking of the fast-switching reference current. The control law was modified accordingly, incorporating \( K_\textrm{ff} = 0.6 \), along with \( K_P = 4.3 \), \( K_I = 1 \), and \( K_D = 1e{-6} \).

The agent's desired action is determined using PPO-clip with a NN consisting of an input layer of size 4 (number of observations), two fully connected hidden layers with 64 units each and Tanh activation, and an output layer distinguishing between the actor and critic. The actor has two units computing logits, with actions sampled from a categorical distribution based on log probabilities. The critic has one unit estimating the value function. Key parameters include a learning rate of \(3 \times 10^{-4}\) for the actor and \(1 \times 10^{-3}\) for the critic, a discount factor \(\gamma = 0.99\), Generalized Advantage Estimation parameter \(\lambda = 0.97\), clipping ratio \(\epsilon = 0.2\), 60 training epochs, 4000 time steps per epoch, a target Kullback–Leibler Divergence (KLD) of 0.01, and optimization using ADAM.

\begin{figure}[H]
    \centering
    \subfigure[Simulation results of Average Episodic Reward (blue star), Average of Episode's Length (red square) and Number of Episodes (green dot) in each epoch.]
    {
        \includegraphics[width=3in,height=1.7in]{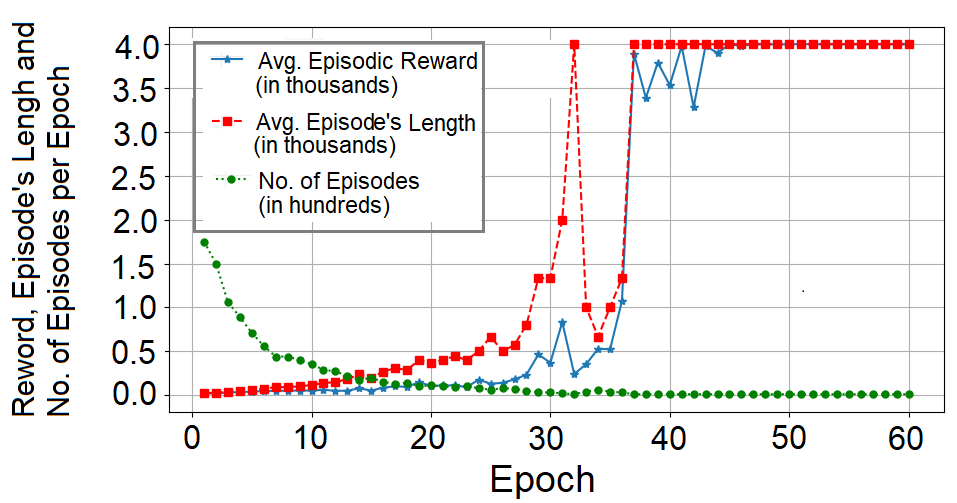}
        \label{Fig Cartpole Reward Kp4.3 Ki1 Kd1e-6 Kff0.6}
    }
    \subfigure[Desired Torque and Actual Torque]
    {
        \includegraphics[width=3in,height=1.65in]{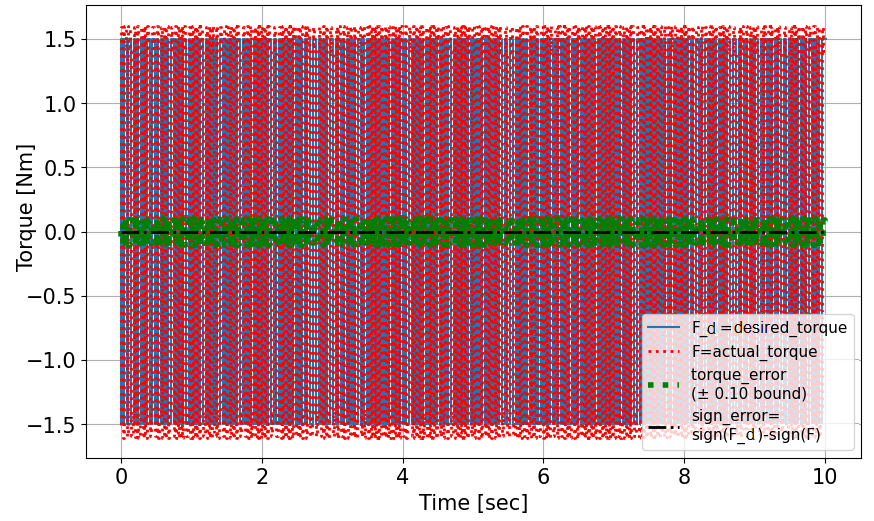}
        
        \label{Fig Cartpole Action Kp4.3 Ki1 Kd1e-6 Kff0.6}
    }
    \subfigure[Pole's angle, $\theta$]
    {
        \includegraphics[width=3in,height=1.7in]{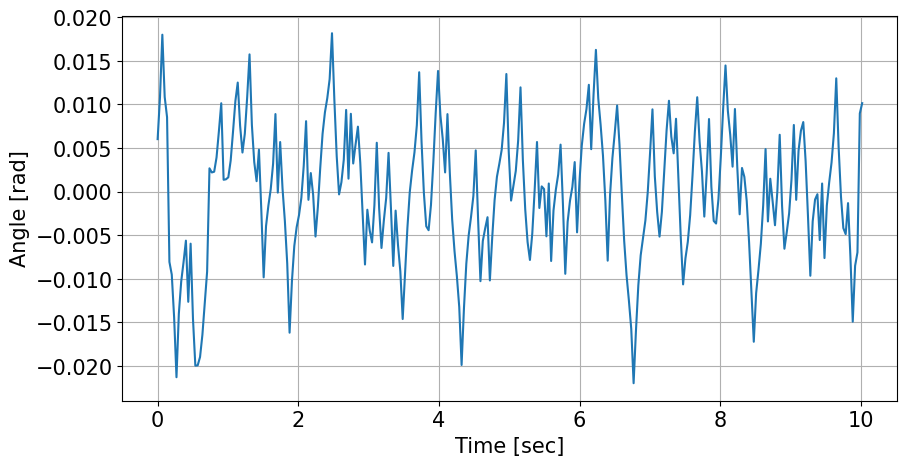}
        \label{Fig Cartpole theta Kp4.3 Ki1 Kd1e-6 Kff0.6}
    }
    \subfigure[Cart's position, $x$]
    {
        \includegraphics[width=3in,height=1.7in]{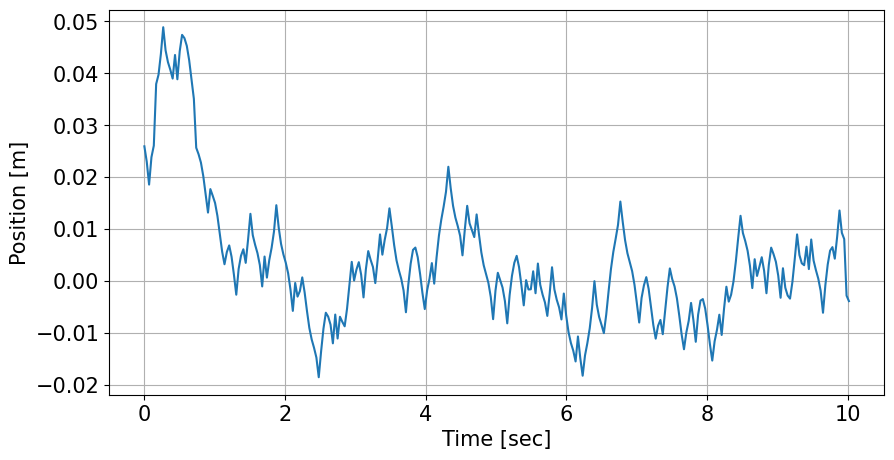}
        \label{Fig Cartpole x Kp4.3 Ki1 Kd1e-6 Kff0.6}
    }
    \caption{Simulation results for the Cartpole environment. Simulation parameters: $K_P$=4.3, $K_I$=1 and $K_D$=1e-6. Additionally, the force-to-torque ratio is 0.15 (torque = 0.15$\times$force), and the control voltage $u_t$ includes an additive feedforward term of $0.6 i^\textrm{ref}_t$.}
    \label{Fig Cartpole Results Kp4.3 Ki1 Kd1e-6 Kff0.6}
\end{figure}

The learning process runs for 60 epochs with up to 4000 time steps per epoch. A new episode begins upon termination (pole falling) or reaching the time limit, whichever comes first, with policy and value function updates occurring within episodes. Each epoch records the number of episodes, average episodic return, and average episode length, which differ due to the penalty imposed for keeping the cart within $[-0.1, 0.1]$. After training, the model was tested for 500 time steps (10 seconds) or until termination, with results presented in Fig. \ref{Fig Cartpole Results Kp4.3 Ki1 Kd1e-6 Kff0.6}.

From the learning graphs of average episodic reward, episode length, and number of episodes per epoch (Fig. \ref{Fig Cartpole Reward Kp4.3 Ki1 Kd1e-6 Kff0.6}), we observe that initially, there are many episodes as the agent tries to keep the pole upright, with higher rewards and episode lengths. As training progresses, the reward and episode length averages increase up to the 4000 step limit, while the number of episodes decreases to 1. This indicates that the agent has kept the pole upright for the full epoch, though the average reward may not reach 4000. For example, in epoch 32, the episode length jumps to 4000, but the reward averages 240 due to occasional zero rewards when the cart is near the edges of the allowed region. After epoch 35, the average reward approaches 4000 within a single episode. In the action graph (Fig. \ref{Fig Cartpole Action Kp4.3 Ki1 Kd1e-6 Kff0.6}), the actual torque (red) closely tracks the desired torque (blue), with the error (green) bounded within $\pm$0.1 [Nm]. From the graphs of the pole's angle (Fig. \ref{Fig Cartpole theta Kp4.3 Ki1 Kd1e-6 Kff0.6}) and the cart's position (Fig. \ref{Fig Cartpole x Kp4.3 Ki1 Kd1e-6 Kff0.6}), we see that the agent successfully kept the cartpole within the required bounds (cart position between [-2.4, 2.4] and pole angle between [-0.2095, 0.2095]). Additionally, through our reward shaping (penalizing the agent with zero reward for cart positions outside [-0.1, 0.1]), the agent was encouraged to keep the cart near the origin rather than an arbitrary point within the range.

\section{Conclusion}
\label{sec:conclusion}

In this research, we introduced a novel DRL framework that incorporates control theory to address uncertainties in action execution. By distinguishing between the Planner, which determines the desired action, and the Controller, which selects the control signal to align execution, our approach optimizes both processes in a unified manner.

Through simulations in various mechanical environments, we demonstrated the framework's robustness in handling execution uncertainties. Integrating control mechanisms into DRL enhances adaptability and performance, particularly in automation and robotics, where precise execution is crucial. Our results highlight the importance of addressing real-world uncertainties to improve the reliability of AI-driven decision-making systems.

Our findings bridge the gap between high-level decision-making and low-level execution, offering a practical solution for deploying DRL in real-world applications. The open-source implementation provides a valuable resource for practical adoption, and this work has the potential to contribute to the broader success of DRL in automation and control. By enhancing robustness and adaptability, this approach paves the way for more effective deployment of AI-driven systems in dynamic and uncertain environments.

\section{Appendix}

\subsection{Implementation Details of the Open Source Software}

This appendix summarizes the implementation details of our open-source Python software (\cite{web:OrenGit}). For each Gym classic control environment, we provide a Python notebook implementing feedback control on the desired action using a PID controller and a DC motor model (see Fig. \ref{Fig GitHub1} for the list of notebooks). Each notebook includes:

\begin{figure}[H]
    \centering
    \includegraphics[width=3.5in,height=3in]{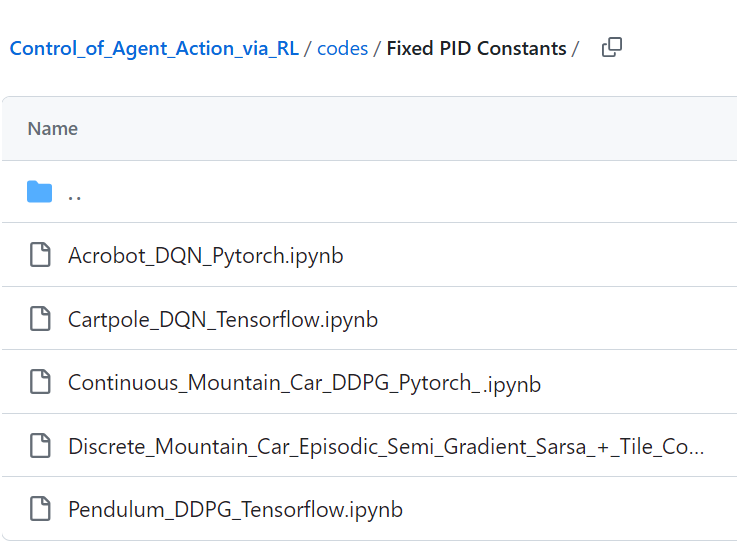}
    \caption{The project's directory of the feedback control of the desired action in GitHub \cite{web:OrenGit}} 
    \label{Fig GitHub1}
\end{figure}

\noindent 
1) A customized version of Gym’s classic control environment class, incorporating a DC motor model. This model calculates the actual force/torque applied based on the desired input, the actuated body’s velocity, and a PID-controlled DC motor field. Users can modify system parameters or adjust reward shaping.

\noindent
2) A new \texttt{ElectricalDCMotorEnv()} class, defining DC motor parameters (e.g., resistor, inductor, torque and voltage constants) and PID control settings. It also offers optional approximations for integral and derivative components via low-pass and high-pass filters. The motor model updates electrical current at each step and outputs the applied force/torque.

\noindent
3) A DRL implementation for selecting the desired action. For instance, we use DDPG in TensorFlow for the Pendulum environment, but users can employ different frameworks (e.g., PyTorch) or test custom DRL algorithms.

\noindent
4) Training and test results, showing accumulated rewards over episodes and comparisons of desired vs. actual force/torque, along with state/observation graphs.

\bibliographystyle{ieeetr}


\end{document}